\documentclass[10pt,letterpaper]{article}
\usepackage[top=0.85in,left=2.75in,footskip=0.75in]{geometry}

\usepackage{amsmath,amssymb}

\usepackage{changepage}

\usepackage{textcomp,marvosym}
\usepackage{booktabs}

\usepackage{cite}

\usepackage{nameref,hyperref}

\UseRawInputEncoding
\usepackage[nopatch=eqnum]{microtype}
\DisableLigatures[f]{encoding = *, family = * }

\usepackage[table]{xcolor}

\usepackage{array}
\usepackage{float}
\usepackage{longtable}
\usepackage{subcaption}

\newcolumntype{+}{!{\vrule width 2pt}}

\newlength\savedwidth

\raggedright
\usepackage[aboveskip=1pt,labelfont=bf,labelsep=period,justification=raggedright,singlelinecheck=off]{caption}

\makeatletter
\renewcommand{\@biblabel}[1]{\quad#1.}
\makeatother

\usepackage{lastpage,fancyhdr,graphicx}
\usepackage{epstopdf}
\fancyheadoffset[L]{2.25in}
\fancyfootoffset[L]{2.25in}
\begin{document}

\vspace*{0.2in}

\begin{flushleft}
{\Large
\textbf\newline{Academic case reports lack diversity: Assessing the presence and diversity of sociodemographic and behavioral factors related to Post COVID-19 Condition} 
}
\newline
\\
Juan Andres Medina Florez\textsuperscript{1},
Shaina Raza\textsuperscript{3},
Rashida Lynn\textsuperscript{2},
Zahra Shakeri\textsuperscript{1},
Brendan T. Smith\textsuperscript{1,4},
Elham Dolatabadi\textsuperscript{1,2,3*}
\\
\bigskip
\textbf{1} Dalla Lana School of Public Health, University of Toronto, Canada
\\
\textbf{2} York University, Canada
\\
\textbf{3} Vector Institute, Canada
\\
\textbf{4} Public Health Ontario, Canada
\\
\bigskip

%
%

* edolatab@yorku.ca

\end{flushleft}
\section*{Abstract}
Understanding disparities in the prevalence of Post COVID-19 Condition (PCC) amongst vulnerable populations is crucial to improving care and addressing intersecting inequities. This study aims to develop a comprehensive framework for integrating social determinants of health (SDOH) into PCC research by leveraging natural language processing (NLP) techniques to analyze disparities and variations in SDOH representation within PCC case reports. Following construction of a PCC Case Report Corpus, comprising over 7,000 case reports from the LitCOVID repository, a subset of 709 reports were annotated with 26 core SDOH-related entity types using pre-trained named entity recognition (NER) models, human review, and data augmentation to improve quality, diversity and representation of entity types. An NLP pipeline integrating NER, natural language inference (NLI), trigram and frequency analyses was developed to extract and analyze these entities. Both encoder-only transformer models and RNN-based models were assessed for the NER objective.\\
Fine-tuned encoder-only BERT models outperformed traditional RNN-based models in generalizability to distinct sentence structures and greater class sparsity. Exploratory analysis revealed variability in entity richness, with prevalent entities like condition, age, and access to care, and underrepresentation of sensitive categories like race and housing status. Trigram analysis highlighted frequent co-occurrences among entities, including age, gender, and condition. The NLI objective (entailment and contradiction analysis) showed attributes like "Experienced violence or abuse" and "Has medical insurance" had high entailment rates (82.4\%-80.3\%), while attributes such as "Is female-identifying," "Is married," and "Has a terminal condition" exhibited high contradiction rates (70.8\%-98.5\%).\\
We conclude that transformer-based NER models are effective in extracting SDOH information from PCC case reports; however, there is a scarcity of multiple sociodemographic factors in PCC-related academic case reports and an imbalanced representation of mentions aligned with groups of interest within dimensions such as gender, insurance status, and age.

\section*{Introduction}
The acute phase of the COVID-19 pandemic may have passed, but its long-term impacts continue to affect millions worldwide \cite{Davis2021-ey,Davis2023-rv,Dolatabadi2023-ex}. According to the World Health Organization (WHO) \cite{noauthor_undated-fk}, 10–20\% of COVID-19 cases lead to post-COVID condition (PCC), also known as long COVID. The Centers for Disease Control and Prevention (CDC) \cite{Cdc2024-wb} reports 6.9\% of U.S. adults have experienced PCC, with 3.4\% still symptomatic. A November 2024 analysis by United Press International (UPI) \cite{noauthor_2024-ti} estimates that 23\% of Americans who had COVID-19 may experience symptoms of PCC, highlighting the substantial and ongoing public health challenges posed by PCC. In response to the pressing need for timely mitigation of PCC, various research initiatives have been deployed and strengthened in the years following the peak of the COVID-19 pandemic. These efforts range from addressing the difficulties experienced by those living with PCC to the characterization of PCC phenomenology to quality of care improvement, and the identification of the biological, social and behavioral determinants associated with PCC development \cite{Bergmans2024-kx,Michelen2021-rm,Sadat_Larijani2022-rj,Greenhalgh2024-yf,Mateu2023-nj}. Sociodemographic and behavioral determinants, such as race, income, and occupational status, are critical for understanding how structural social factors result in disparities in PCC prevalence, severity, and access to care \cite{Raza2023-ih,Al-Aly2024-pp,Egede2024-qv}. While studies focusing on the social and behavioral determinants of PCC remain limited, existing research highlights the disproportionate prevalence of PCC among certain population subgroups, including females, transgender individuals, and Hispanics \cite{Bai2022-wn,Fernandez-de-Las-Penas2022-vt,Louie2023-rx,Thomason2022-hy,Jacobs2023-tk}. Importantly, as scientific knowledge of PCC continues to expand and deepen \cite{noauthor_2022-ew}, a significant knowledge gap persists concerning why disparities in PCC prevalence amongst subgroups exist or how to reduce them.

The lack of comprehensive datasets that include both sociodemographic and behavioral variables, such as race, ethnicity, age, income, and occupational status, and PCC is a prominent barrier \cite{Khunti2024-wf,Prashar2023-fj}. For instance, structured clinical Datasets, such as Electronic or Medical Health Records often fail to include detailed sociodemographic variables, limiting analyses of how systemic inequities impact PCC outcomes, such as unequal access to healthcare and its impact on PCC outcomes \cite{Banerjee2024-dd,Hua2024-gd,Emani2023-pi,Cook2022-kz}. Research has highlighted the inadequacy of race and ethnicity data in electronic health records, an issue compounded by inconsistent collection and reporting practices \cite{Cook2022-kz,noauthor_undated-ll,Staroselsky2006-fk}. The scarcity of such data hinders efforts to identify actionable pathways for disparity reduction. Bridging these gaps requires advancements in data collection and integration. Furthermore, the integration of additional data sources, such as clinical case reports, could complement structured datasets, offering richer insights into patients' experiences and social contexts. By combining these reports with advanced natural language processing (NLP) techniques—such as entity extraction \cite{Richie2023-mr}, contextual understanding \cite{Yang2022-di}, language entailment \cite{Seerat2024-ul}, and classification \cite{Dolatabadi2023-wr}—researchers can uncover complex relationships between sociodemographic factors and PCC development \cite{Patra2021-yy,Bompelli2021-rb}.

The overarching objective of this study is to establish an approach for advancing the integration of SDOH into PCC research through state-of-the-art NLP methodologies. Specifically, this study aims to develop and apply a transformer-based NER model to automate the extraction of sociodemographic and behavioral determinants of health, analyze their representation and diversity within the PCC academic literature, and identify disparities and gaps in this domain. By doing so, this study seeks to inform improvements in data comprehensiveness and equity, ultimately guiding more inclusive and effective research and policy development for PCC. This study makes three contributions to the field:

\begin{enumerate}
    \item We release an open-source, novel PCC Case Report Corpus comprising more than 7,000 case reports, annotated using the proposed fine-tuned BERT-base-uncased model. This corpus includes 26 sociodemographic, behavioral, and clinical entities such as age, vaccination status and medical condition, along with a robust data processing pipeline designed to improve reproducibility and facilitate future research in PCC. We are releasing our dataset to the research community for reproducibility of the experiments. The dataset is hosted on Hugging Face and is accessible via the following link: \url{https://huggingface.co/datasets/PCC-SDOH-NLP/longCOVID_CaseReports_SDOH}.
    
    \item We introduce a comprehensive end-to-end NLP pipeline that integrates named entity recognition (NER) and natural language inference (NLI) techniques, enabling the extraction and entailment of entities from case reports. This pipeline utilizes a combination of data augmentation, regularization, rule-based methods and generative AI, and categorizes the extracted entities into meaningful groupings, providing a structured framework for further analysis.
    \item We identify key gaps in the representation of SDOH attributes in PCC, such as underrepresentation of race and spiritual beliefs, and reveal patterns of agreement and contradictions in entity co-occurrences, such as disparities between documented insurance coverage and access to care. 
\end{enumerate}

\subsection*{Related Work}
While the utilization of NLP in the context of PCC is not novel, these efforts have been largely focused on the identification of clinical and biological entities, such as symptoms, diagnoses, and treatments \cite{Dolatabadi2023-ex,Patel2023-gi,Han2023-oe,Bhambhoria2023-vl,Koss2022-xu,Zhu2022-mv}. To our knowledge, NLP techniques have not been specifically tailored to extract a comprehensive set of sociodemographic entities in the PCC context.

\section*{Materials and methods}
\subsection*{Corpus Construction for PCC Case Reports}
The PCC Case Report Corpus was developed by sourcing relevant case report articles from the LitCOVID repository \cite{Chen2023-hl} guided by a query incorporating the keywords "Post COVID," "Long COVID," and "Post-acute COVID-19 syndrome." Inclusion criteria specified case reports published from January 1, 2020, to October 16, 2023, in English, featuring patients with a confirmed history of SARS-CoV-2 infection and documented PCC symptoms or complications. Only full-text articles were considered. Exclusion criteria included preprints, non peer-reviewed articles, review articles, meta-analyses, systematic reviews, studies focused exclusively on acute COVID-19, articles lacking detailed clinical information, and non-human studies.

\begin{figure}[H]
    \centering
    \includegraphics[width=1\linewidth]{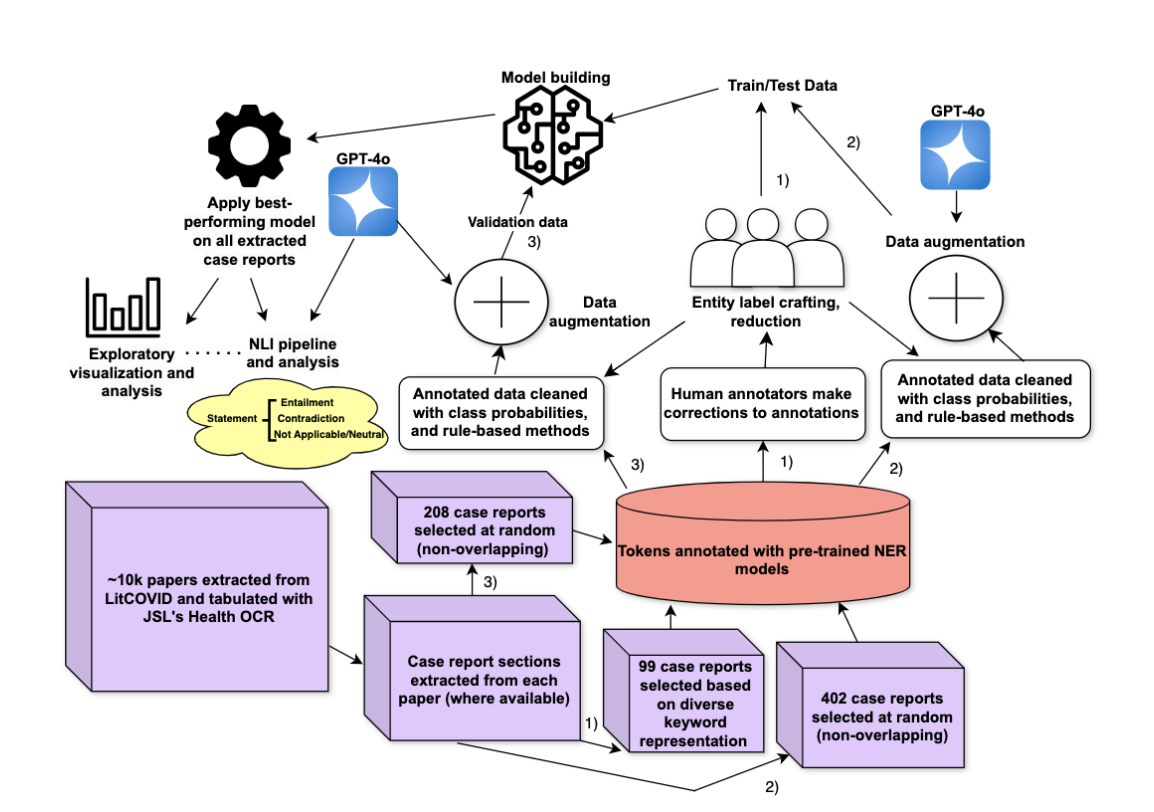}
    
    \caption{The PCC Case Report Corpus was developed from LitCOVID case reports (January 2020–October 2023) featuring post-COVID symptoms, using keyword-based searches and strict inclusion criteria. After filtering and processing, 7,172 case reports were curated, with 709 selected for NER model development: 99 in subset 1 (keyword-filtered for diversity), 402 in subset 2, and 208 in subset 3 (randomly selected). Annotation involved initial NER, human review, and data augmentation, balancing quality and efficiency to enhance the model's ability to capture PCC complexities.}
    \label{fig: 1}
\end{figure}

Approximately 10,000 papers meeting these criteria were retrieved from LitCOVID. PDFs of the relevant articles were converted into images, processed with optical character recognition (OCR), and structured using the John Snow Labs Health OCR tool \cite{noauthor_undated-fd} (see Fig \ref{fig: 1} for the full pipeline). To minimize background noise, the case report sections were extracted from each document, as preliminary exploration showed that these sections contained most of the socioeconomic and clinical attributes relevant to the subjects of interest. Non-relevant content was systematically excluded using rule-based regular expressions, such that documents lacking case report sections were removed, and rules were iteratively refined based on observed errors. This process yielded a curated corpus of 7,172 case reports for further analysis. Three non-overlapping subsets, totaling 709 case reports, were selected for NER model development. These are referred to as subset 1, with 99 case reports, subset 2, with 402 case reports, and subset 3, with 208 case reports, as shown in Fig \ref{fig: 1}. Of these, subsets 1 and 2, totaling 501 case reports, were used for model training, with an 80/20 random split applied to create a training set and an optimization testing set. The remaining 208 case reports (subset 3) were reserved for evaluating model generalization. The size of each subset was determined based on the feasibility of processing and annotation, while ensuring that the sizes remained relatively proportionate to each other based on their end purpose (development vs. validation set). In this study, the 20\% subset from the training data is referred to as the optimization testing set, while the 208 reserved case reports are referred to as the generalization evaluation set.\\

The 99 case reports utilized for subset 1 were selected with a keyword-based and rule-based filtration process to ensure that there would be a representation of a variety of labels and that overall text quality would be adequate. In particular, keywords representing marginalized and at-risk populations were prioritized to ensure diversity in our training data, including homelessness, housing-insecurity, low-income, poverty, black, hispanic, uninsured, abused, bisexual, homosexual, and female. Subsets 2 and 3, on the other hand, were selected at random from non-overlapping subsets of the dataset. Additional initial explorations on the corpus are outlined in supplementary materials.
\newpage
\subsection*{Annotation Process}

The annotation process consisted of four main steps of initial entity extraction using pre-trained NER models by John Snow Labs (JSL), human review, entity refinement and a combination of data filtration and augmentation. The entire dataset of 709 case reports underwent the initial entity extraction step, while steps of human review, and data filtration and augmentation were applied to targeted subsets 1 and 2-3, respectively. This tiered approach addressed two key challenges: first, that full manual annotation of all 709 reports would be resource-intensive, and second, that the use of a smaller subset for human annotation allowed for subsequent automated filtration and augmentation to be explored on the remaining reports. By integrating human-verified labels with additional, augmented data, this approach maximized both data quality and diversity, enabling the model to better capture the complexity of PCC literature.

\subsubsection*{Step 1- Initial Entity Extraction Using Pre-Trained NER Models}
Two pre-trained NER models from JSL were applied in parallel to the full set of model development case reports to generate preliminary entity annotations. The first model, “ner\_sdoh” \cite{noauthor_2023-qw}, was trained to identify 47 entities spanning social and clinical factors, while the second model, “ner\_covid\_trials” \cite{noauthor_2021-mb}, was trained to categorize 41 entities including attributes relevant to COVID-19. Based on the alignment of “ner\_sdoh” with our proposed objective, “ner\_sdoh” was utilized as the base model and annotations from “ner\_covid\_trials” replaced those from “ner\_sdoh” for certain critical entities. Namely, initial explorations revealed a superior performance of “ner\_covid\_trials” in the age entity. Vaccination status, which was not included in “ner\_sdoh,” was also included due to its relevance to PCC. This substitution enhanced dataset relevance by ensuring that key entities were more accurately represented.
\subsubsection*{Step 2 - Human Annotation}
Following initial extraction, subset 1, comprising 99 case reports, was selected to undergo human annotation. This subset was curated to include adequate representation of  a variety of entities of interest, with particular emphasis on keywords related to marginalized and at-risk populations to ensure diverse representation in the training data. The subset was divided equally among three human annotators tasked with reviewing and refining the model-predicted labels. Before annotation, the annotators reached a consensus on entity definitions and labeling criteria, which was essential for consistency and reduced inter-annotator variability. This preparatory step ensured uniformity in label interpretation, thus enhancing the reliability and precision of the final annotated dataset.
\subsubsection*{Step 3: Entity Type Refinement}
Following expert advice, a review of the WHO’s SDOH framework \cite{noauthor_2010-nr}, and under the guidance of three human annotators with health-related experience, we implemented an entity type refinement process. This was aimed at enhancing our model’s classification accuracy while preserving critical sociodemographic dimensions relevant to our study. Importantly, this process entailed combining highly interrelated labels and eliminating entities outside of the research focus of this analysis. The entity label dimensions included, excluded and refined are summarized in Table \ref{tab:1}.

The base models from JSL initially yielded over 45 distinct entities (resulting in 90 labels in CoNLL format), of which 14 (28 in CoNLL format) were identified as outside of the scope of this analysis. More details on the refinement process are outlined in supplementary materials (sFigure 1). This refinement resulted in a set of 27 core entity types. For CoNLL format consistency, each label (except for 'O') was divided into 'B-' (beginning) and 'I-' (inside) sub-labels, resulting in a total of 53 distinct types for model training.

\begin{table}[h]
    \centering
    \fontsize{8}{10}\selectfont  
    \rmfamily
    \begin{tabular}{|p{3.5cm}|p{3.5cm}|p{3.5cm}|}  
        \hline
        \textbf{Entity Label Dimensions Excluded (Classified as ‘O’) from Pretrained Models} &
        \textbf{Entity Label Dimensions Included from Pretrained Models} &
        \textbf{Refined Entity Labels} \\ \hline
        \parbox{5cm}{\texttt{"Community Safety"} \\ \texttt{"Date"} \\ \texttt{"Healthcare Institution"} \\ \texttt{"Legal Issues"} \\ \texttt{"Other SDoH Keywords"} \\ \texttt{"Population Group"} \\ \texttt{"Quality Of Life"} \\ \texttt{"Sexual Activity"} \\ \texttt{"Substance Duration"} \\ \texttt{"Substance Frequency"} \\ \texttt{"Substance Quantity"} \\ \texttt{"Transportation"} \\ \texttt{"Childhood Event"} \\ \texttt{"Environmental Condition"}} &
        \parbox{3.5cm}{\texttt{"Access To Care"} \\ \texttt{"Age"} \\ \texttt{"Alcohol"} \\ \texttt{"Communicable Disease"} \\ \texttt{"Diet"} \\ \texttt{"Disability"} \\ \texttt{"Eating Disorder"} \\ \texttt{"Education"} \\ \texttt{"Employment"} \\ \texttt{"Exercise"} \\ \texttt{"Family Member"} \\ \texttt{"Financial Status"} \\ \texttt{"Food Insecurity"} \\ \texttt{"Gender"} \\ \texttt{"Geographic Entity"} \\ \texttt{"Housing"} \\ \texttt{"Hyperlipidemia"} \\ \texttt{"Hypertension"} \\ \texttt{"Income"} \\ \texttt{"Insurance Status"} \\ \texttt{"Language"} \\ \texttt{"Marital Status"} \\ \texttt{"Mental Health"} \\ \texttt{"Obesity"} \\ \texttt{"Other Disease"} \\ \texttt{"Race Ethnicity"} \\ \texttt{"Sexual Orientation"} \\ \texttt{"Smoking"} \\ \texttt{"Social Exclusion"} \\ \texttt{"Social Support"} \\ \texttt{"Spiritual Beliefs"} \\ \texttt{"Substance Use"} \\ \texttt{"Transportation"} \\ \texttt{"Violence Or Abuse"} \\ \texttt{"Vaccine"} \\ \texttt{"Admission Discharge"}} &
        \parbox{3.5cm}{\texttt{"Access To Care"} \\ \texttt{"Age"} \\ \texttt{"Condition"} \\ \texttt{"Diet"} \\ \texttt{"Disability"} \\ \texttt{"Education"} \\ \texttt{"Employment"} \\ \texttt{"Exercise"} \\ \texttt{"Family Member"} \\ \texttt{"Gender"} \\ \texttt{"Geographic Entity"} \\ \texttt{"Housing"} \\ \texttt{"Income"} \\ \texttt{"Insurance Status"} \\ \texttt{"Language"} \\ \texttt{"Marital Status"} \\ \texttt{"Mental Health"} \\ \texttt{"Race Ethnicity"} \\ \texttt{"Severity"} \\ \texttt{"Sexual Orientation"} \\ \texttt{"Social Support"} \\ \texttt{"Spiritual Beliefs"} \\ \texttt{"Substance"} \\ \texttt{"Treatment"} \\ \texttt{"Vaccine"} \\ \texttt{"Violence Or Abuse"} \\ \texttt{"O"}} \\
        \hline
    \end{tabular}
    \caption{Entity label refinement process.}
    \label{tab:1}
\end{table}

\newpage
\subsubsection*{Step 4 - Filtration and Data Augmentation}
Addressing the class imbalance, particularly the over-representation of the non-entity (‘O’) class, was essential for improving entity extraction accuracy. To tackle this, a two-stage filtration and augmentation strategy was implemented on subsets 2 and 3, aimed at enhancing the representation of rare entities while mitigating class imbalance. For the first stage, entity types with a probability below 90\% were first reassigned to the 'O' class for added robustness. Subsequently, the entity types were reduced and refined following the process outlined in step 3. In addition, to reduce misclassification noise from numeric values, any numeric labels not identified as 'Age' (either 'B-Age' or 'I-Age') were also converted to 'O'. To further address the prevalence of the no-entity (‘O’) class during training, sentences containing only 'O' class labels were removed from subset 2 alone, while subset 3, which is utilized for evaluation, was not subjected to this additional sentence-level downsampling technique.\\

The augmentation process was applied separately to subset 2, designated for training, and subset 3, designated for evaluation. For the training subset, three sentence templates were generated with GPT-4o \cite{noauthor_undated-kh} to minimize the presence of the 'O' class while retaining all entity types of interest. Both sentence templates and variations were manually reviewed for potential biases, and a manual review of a subset of the produced sentences ensured that there was not a disproportionate prevalence of a particular variation. These templates were tokenized, labeled, and stored in three base dataframes. From these structures, 3,000 sets of synthetic sentences were created by randomly combining rare (non-‘O’) entities, with 1,500 derived from a complete entity variation dictionary (composed of the set of rare entities in the corpus and variations generated by GPT-4o) and 1,500 derived from the variations generated by GPT-4o alone. Each synthetic sentence set was assigned a unique identifier and was randomly embedded within the case reports, seamlessly integrating with the dataset for model training. This augmented 402 case reports was then merged with the 99 human-reviewed case reports to form the final training dataset for the NER model training. To assess potential model overfitting and improve generalizability, the augmentation of subset 3 was designed with greater class sparsity and distinct sentence structures (i.e., different entity type orders) than those used in training. GPT-4 generated three distinct sentence templates, featuring higher ‘O’ class representation and varied label sequences. Importantly, gender was not included in all sentence structures given its greater presence pre-augmentation. As before, each sentence structure was tokenized, labeled, and structured into three base dataframes. Using these dataframes, 3,000 augmented sets of sentences were generated, with 1,500 sentences derived from a complete entity variation dictionary (composed of the set of rare entities in the corpus and variations generated by GPT-4o) and with 1,500 derived from the variations generated by GPT-4o alone. Each synthetic sentence was assigned a unique identifier and was embedded randomly into the evaluation case reports to evaluate model performance. The entity variations produced by GPT-4o and sample prompts utilized to produce the entity type variations and sentence structures are outlined in supplementary materials.
\subsection*{NER Model Development}
\subsubsection*{NER Models}
In this study, both transformer-based models and traditional deep learning architectures are benchmarked to assess their capabilities for the selected NER task, with a particular emphasis on handling imbalanced entity classes and generalizing to unseen textual structures. The primary models of interest were BERT-based, encoder-only transformer models, due to their encoder architecture, which is especially suited for NER tasks due to its transformer encoder-only architecture \cite{Luoma2020-rq}. In particular, BERT-base-uncased \cite{Devlin2018-st,noauthor_undated-lp} and DistilBERT-base-uncased \cite{Sanh2019-me} were selected due to their substantial pre-training on English language texts. In addition, the Biomed-NLP model, which utilizes the BERT architecture, was included in the model exploration due to its pre-training on a large body of biomedical texts \cite{Gu2020-pu}. As baselines, Bidirectional LSTM (BiLSTM), Recurrent Neural Network (RNN), and Gated Recurrent Unit (GRU) models are included due to their ability to process sequential dependencies \cite{Schuster1997-wa,Hopfield1982-bh,Cho2014-hg}.\\

For NER model building, to accommodate the 512-token limit per textual sample, each full string (case report text, plus augmentation in selected cases) was split into segments of up to 512 tokens while ensuring the preservation of sentence boundaries, to minimize contextual loss. For the BERT-based models, 512-token sequences were prepared with CLS, SEP, and PAD tokens, along with a mask layer. Padded sequences were generated for the BiLSTM, RNN, and GRU models.

All BERT models were fine-tuned for 300 epochs with a learning rate of 1e-5. Other deep learning models were fully trained from random initialization for 10 epochs with a learning rate of 1e-9. All models employed a sparse cross-entropy loss function and were optimized using the Adam optimizer, with regularization techniques such as class weighting, sample weighting, and dropout layers to mitigate overfitting on majority classes. Importantly, a lower learning rate was implemented for the training of non-transformer models due to an observed propensity to overfit on the training data during early experiments. Conversely, to optimize the performance of transformer models, such that their performance would more closely assimilate that of non-transformer models on the test set, a higher number of epochs was implemented during training.

For model assessment, the macro F1 score is utilized as the primary performance metric. To account for the influence of the ‘O’ class performance in the macro F1 score, the macro F1 score excluding the ‘O’ class is also assessed. Other key performance metrics assessed include the macro One vs. One (OVO) Area Under the Receiver Operating Characteristic Curve (AUC)  and the macro One vs. Rest (OVR) AUC scores. For the best-performing model, we also present class-specific F1 scores.
\subsection*{NLI Pipeline Development}
To gain more detailed insights into the extracted entities and their corresponding types, a rule-based NLI pipeline was leveraged. Importantly, the entities were utilized as the strings to be assessed, and statements were crafted by entity type to assess entailment, contradiction and neutrality/non-applicability of the entities. These statements are outlined in the supplementary materials. Given the significant variation in meaning and scope among our entity types, the selected statements encompass both the presence and absence of attributes, as well as other meaningful binary distinctions in the data. To support this analysis, we generated lists of words indicating ‘entailment’ and ‘contradiction’ for each entity using prompts in GPT-4. We further enhanced these lists by augmenting them with synonyms from the NLTK library \cite{noauthor_undated-vn}. Using RegEx each extracted entity was matched against the generated lists, returning ‘Entailment,’ ‘Contradiction,’ or ‘Not Applicable’ as appropriate. Importantly, only tokens marked with ‘Entailment’ or ‘Contradiction’ are considered, given that those marked as ‘Not Applicable’ are prone to contain noise and other variations out of the scope of this pipeline. Manual validation on a subset of matches was performed to assess RegEx accuracy and refine patterns. The entailment and contradiction sets (produced by GPT-4o), and the corresponding prompt, are outlined in the supplementary materials.
\subsubsection*{Statement selection}
For entity types including ‘Social\_Support,’ ‘Substance,’ ‘Marital\_Status,’ ‘Disability,’ ‘Housing,’ ‘Insurance\_Status,’ ‘Violence\_Or\_Abuse,’ ‘Employment,’ ‘Vaccine,’ ‘Mental\_Health,’ and ‘Access\_To\_Care.’, the statements indicate the presence or absence of attributes.

For more complex labels representing multiple categories or continuums, the statements focus on insights related to sensitive or at-risk groups. or to highlight other relevant binary breaks in the data. For ‘Education,’ the statement was crafted to create a break between terms specifying a higher formal education level (high school or above) from those that do not. Importantly, having a general education development (GED) diploma has been associated with better health outcomes later in life \cite{Caputo2005-lh}. For ‘Race\_Ethnicity,’ the statement generates a binary break between white/caucasian and non-white, which is significant given the mechanisms of structural racism and their effects on the health experience and outcome of non-white individuals \cite{Brandt2023-dh}. We recognize that this binary classification is limited, as it may not capture intersectional nuances; however, we justify this choice based on the goals of the study to provide initial insights for further study. For ‘Exercise’, the statement is aligned with the documented effect of exercising regularly on the immune response, particularly with regard to COVID-19 \cite{Da_Silveira2021-nb}. For ‘Geographic\_Entity’ and ‘Language,’ the statements were chosen in conjunction to reflect the effect of a language barrier on the health care experience, such that it can be observed if there is a mismatch between the language spoken in the country of origin/care location and the primary language \cite{Venkatesan2022-qw}. Importantly, both statements are dichotomized as ‘English or not’ given that our selected texts are in English, assessing, therefore, if there is an English language dominance in the location and language of subjects in the case reports. For ‘Severity,’ the statement is crafted to identify if the observed symptoms are highly severe or not, which can be of interest to observed potential patterns between long COVID and symptom/condition severity.

In addition, for ‘Diet,’ the statement aims to capture the relationship between dietary restrictions and health, which has been associated with immune responses and other health conditions \cite{Berger2013-ia,Giugliano2006-jc}. For ‘Income,’ the statement was crafted to reflect the effect of higher income level (or lack thereof) on access to health care \cite{Larson2010-gv}. Given that our corpus hails from distinct countries, with distinct thresholds for upper-middle income, we do not include numeric measures in the entailment and contradiction sets. For ‘Family\_Member,’ the statement was crafted to capture the social support provided by progeny, which is an essential form of caretaking in multiple cultural settings, with an associated impact on parental health \cite{Li2022-ey}. For ‘Sexual\_Orietation,’ the statement was selected to align with the discourse of discrimination and barriers to care faced by non-heterosexual individuals \cite{Alencar_Albuquerque2016-ny}. For ‘Gender,’ the statement was selected to reflect the discussed positive associational relationship between female gender, sex on the development of PCC \cite{Bai2022-wn,Fernandez-de-Las-Penas2022-vt}. For ‘Spiritual\_Beliefs,’ the statement was crafted to align with the historical privilege faced by those of catholic and christian belief systems \cite{Blumenfeld2006-gu}, and the lack thereof faced by several other faith systems, which can be reflected in experiences of discrimination in health care settings \cite{Eriksson2023-vq}. Lastly, for ‘Age,’ ‘Treatment,’ ‘Condition’ and ‘Severity’ the statements were crafted to highlight at-risk groups including older adults and other inmuno-compromised groups, such as those with a highly invasive treatment and those with highly severe, rare, chronic or terminal conditions \cite{Kuy2020-mc,Schneider2024-az}.

\section*{Results}
\subsection*{Distribution of Entity Types Across Case Reports}
\begin{figure}[H]
    \centering
    \begin{subfigure}{\textwidth}
        \centering
        \includegraphics[width=\textwidth]{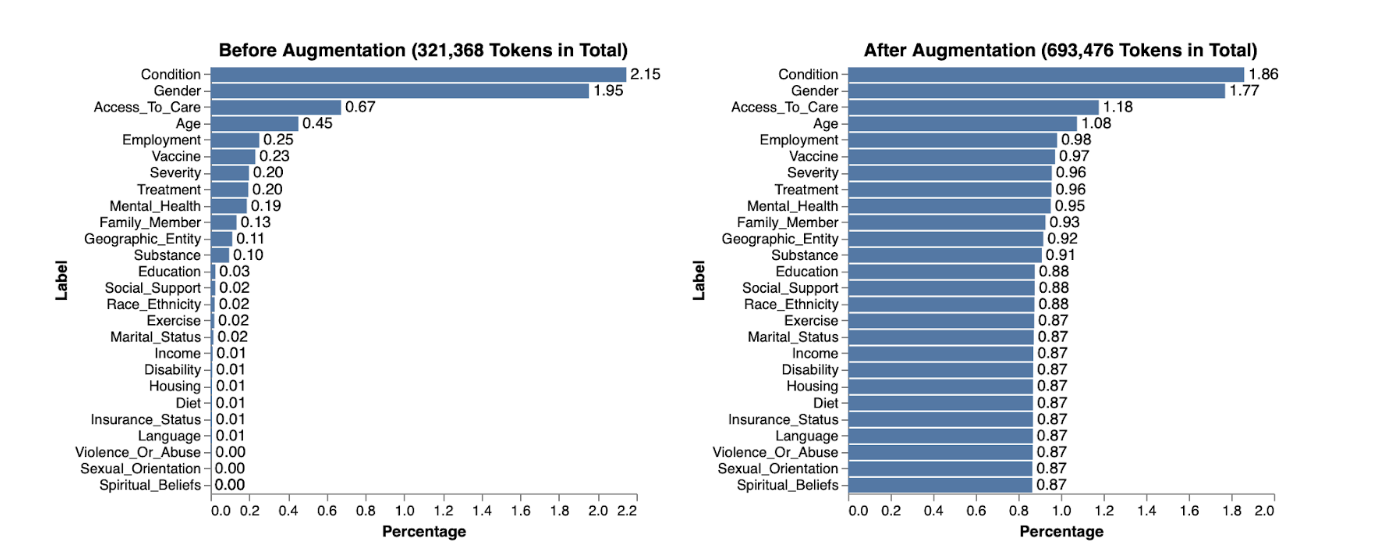}
        \captionsetup{justification=centering}
        \caption{}
    \end{subfigure} \\[1em]  
    \begin{subfigure}{\textwidth}
        \centering
        \includegraphics[width=\textwidth]{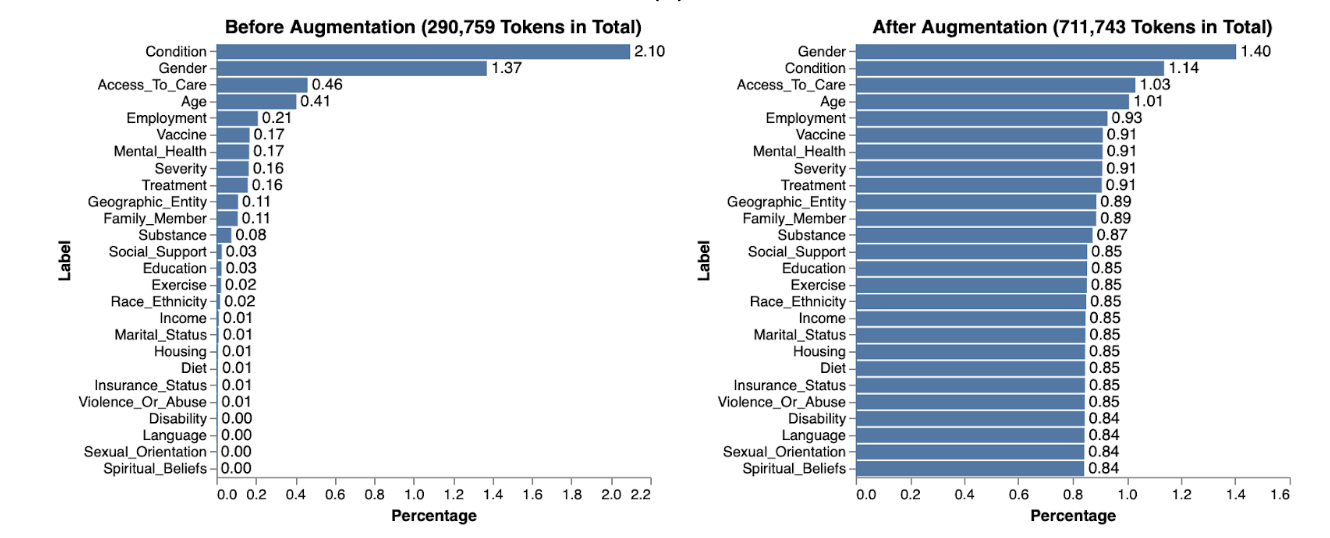}
        \captionsetup{justification=centering}
        \caption{}
    \end{subfigure}
    \caption{Entity frequencies (percentage) for (a) model training and (b) evaluation case reports before and after augmentation. The percentages are calculated out of the total number of tokens, which were 321,368 before augmentation and 693,476 after augmentation for model training and 290,759 before augmentation and 711,743 after augmentation for model evaluation.}  
    \label{fig:2}
\end{figure}

The distribution of all 26 ground truth entity types (excluding the NuLL entity) across the case reports for the model training and generalization evaluation before and after augmentation is illustrated in Fig \ref{fig:2}. Post-augmentation, the model training set and the generalization evaluation set contain 693,476 and 711,743 entities, respectively. The complete collection of case reports contains 1,405,219 entities. ‘Condition’, ‘Gender’, ‘Access to care’, ‘Age’, and ‘Employment’ are the most frequent entities in the 709 case reports leveraged for model training and evaluation. The greater presence of these entity labels reflect biases and oversimplification of SDOH entity dimensions in PCC case reports. As the least prevalent entities, label dimensions such as ‘Spiritual\_Beliefs’ and ‘Sexual\_Orientation’ saw the greatest effect in variational diversity and relative frequency post-augmentation (increasing from 0.0\% to over 0.8\% on the training and generalization evaluation sets).

\subsection*{Comparative Benchmarking Analysis of NER Models}
In benchmarking the RNN-based models, including BiLSTM, BERT, DistilBERT, and BioBERT, we evaluated their performance in extracting 27 entity types on the optimization testing set and generalization evaluation set, as detailed in Table \ref{tab:2}. As the primary  metric of model assessment, macro F1 score, excluding the ‘O’ class, is utilized. Excluding the 'O' class reduces bias from overrepresented non-entity annotations, highlighting the model's ability to accurately classify diverse entity types. The macro-average results for these benchmarks are presented, demonstrating the comparative performance of each model in this task with Macro F1-score of .72 excluding O and macro AUC of .99. We show that BERT models are on average more generalizable than the RNN-based models, although RNN models outperform BERT for the optimization testing set.

The model demonstrated strong performance in classes such as ‘I-Race\_Ethnicity’ and ‘I-Marital\_Status’, both achieving an F1-Score of 0.99, indicating high precision and recall (supplementary materials). Other well-predicted classes included ‘I-Language’ and ‘I-Spiritual\_Beliefs’ (0.98), and ‘B-Education’ and ‘I-Violence\_Or\_Abuse’ (0.95). On the contrary, the least predicted classes had notably lower performance. ‘B-Condition’ and ‘I-Condition’ were the lowest, with F1-scores of 0.17 and 0.14, respectively, indicating difficulties in recognizing specific health conditions. ‘B-Gender’ and classes like ‘B-Diet’ and ‘B-Insurance\_Status’ also underperformed, with scores below 0.40. Exploratory analysis of a subset of classifications in each class revealed that misclassifications stem from model limitations to various health conditions, and to ambiguous mentions and tokenization artifacts.
\begin{table}[h]
    \centering
    \small
    \begin{tabular}{lccccccc}  
        \hline
        \textbf{Model} & \multicolumn{2}{c}{\textbf{Optimization Testing Set}} & \multicolumn{2}{c}{\textbf{Generalization Evaluation Set}} \\ \hline
                        & \textbf{Macro F1-score} & \textbf{Macro AUC} & \textbf{Macro F1-score} & \textbf{Macro AUC} \\ \hline
        BERT            & 0.78 & 0.78 & \textbf{0.72} & \textbf{0.72} \\ 
        DistilBERT      & 0.79 & 0.78 & 0.63 & 0.62 \\ 
        BiomedNLP       & 0.58 & 0.58 & 0.28 & 0.27 \\ 
        BiLSTM          & \textbf{0.95} & \textbf{0.95} & 0.10 & 0.09 \\ \hline
    \end{tabular}
    \caption{The optimization testing set and the generalization evaluation set. The optimization testing set consists of a randomly selected 20 percent subset of the training data (501 case reports and 3000 sets of synthetic sentences), while the generalization evaluation set is an entirely separate subset from the case report cohort, including 208 case reports and 3000 sets of synthetic sentences. OVO denotes One vs. One and OVR denotes One vs. Rest macro AUC scores. BERT is the Base Uncased model.}
    
    \label{tab:2}
\end{table}
\subsection*{Exploratory Analysis of the Entities Extracted by the Best-performing Model}
\begin{figure}[H]
    \centering
    \begin{subfigure}{0.6\textwidth}  
        \centering
        \includegraphics[width=\textwidth]{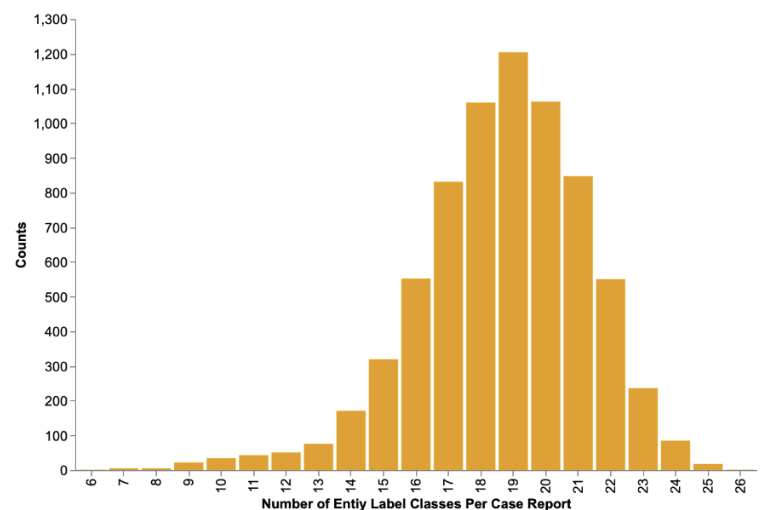}
        \captionsetup{justification=centering}
        \caption{}
    \end{subfigure}
    \hfill
    \begin{subfigure}{0.6\textwidth}  
        \centering
        \includegraphics[width=\textwidth]{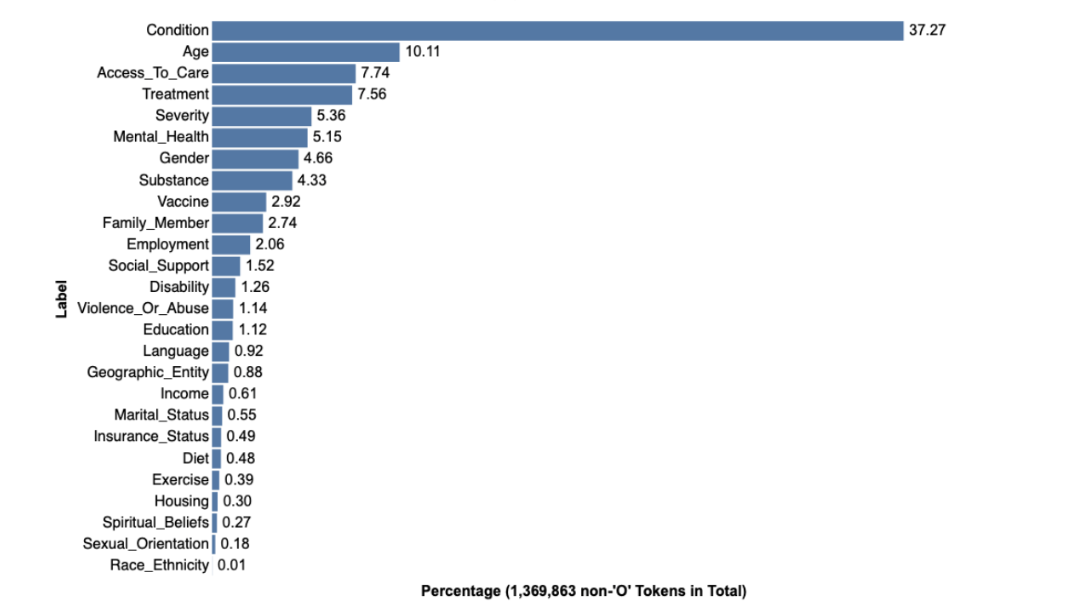}
        \captionsetup{justification=centering}
        \caption{}
    \end{subfigure}
    \hfill
    \begin{subfigure}{0.6\textwidth}  
        \centering
        \includegraphics[width=\textwidth]{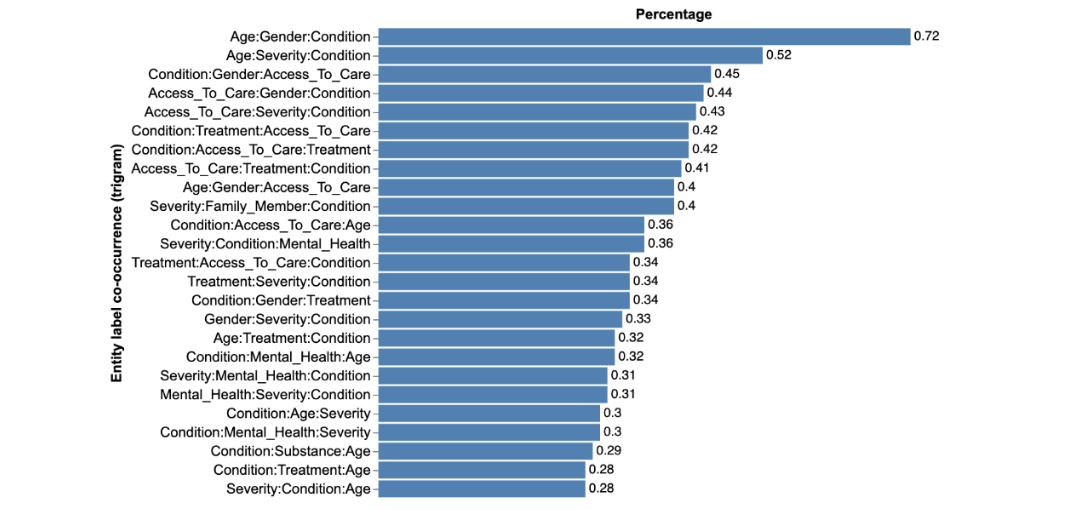}
        \captionsetup{justification=centering}
        \caption{}
    \end{subfigure}

    \caption{Entity extraction and distribution analysis from 7,172 case report sections processed by the fine-tuned BERT Base Uncased model. (a) Distribution of entity counts per case report. (b) Frequency of  non-‘O’ entity labels, out of all non-‘O’ entities. (c) Trigram frequency visualization for the top 25 most frequent entity types, where the order of occurrence is considered.}  
    \label{fig:3}
\end{figure}

The best-performing model, a fine-tuned BERT-base-uncased, was applied to all 7,172 extracted case report sections. A total of 1,369,863 entities were extracted across 26 distinct entity type dimensions, excluding the ‘O’ entity (amounting to 2,097,047 entities alone). The distribution of entity counts per case report is illustrated in Fig \ref{fig:3} (a). We observed variability in entity richness across the corpus, ranging from 6 to 26 distinct non-'O' types. The majority of case reports contained 18 to 20 distinct entity types, with 19 being the most common, observed in 1,204 case reports. A small subset of reports (18) included 25 distinct entity types, while only one case report encompassed all 26 non-'O' entity types.\\

The frequency of non-‘O’ entity types, out of all non-‘O’ entities is depicted in Fig \ref{fig:3} (b). The most frequently mentioned categories include biological components labelled as ‘Condition’ (37.27\%), ‘Age’(10.11\%), ‘Access\_To\_Care’ (7.74\%), ‘Treatment’(7.56\%), ‘Severity’ (5.36\%), ‘Mental\_Health’ (5.15\%), ‘Gender’ (4.66\%), and ‘Substance’ (4.33\%). Conversely, entities such as ‘sexual orientation’ (0.18\%), ‘spiritual beliefs’ (0.27\%), ‘Race\_Ethnicity’ (0.01\%), and ‘housing status’ (0.30\%) are among the least represented, aligning with their sensitive nature.\\

Among the top 25 most frequently occurring entities, we visualized the frequencies of three-entity sequences (trigrams) in Figure 2c, where the order of occurrence is considered. Among the most frequent entity type trigrams, ‘Age,Gender,Condition’ stands out with a frequency of 0.72\%, followed by ‘Age,Severity,Condition’ at 0.52\% and ‘Condition:Gender:Access\_To\_Care’ at 0.45\%. Other notable trigrams include combinations of ‘Access\_To\_Care’,’Gender’,’Condition’, ‘Severity’, ‘Treatment’, ‘Age’, ‘Family\_Member’, ‘Mental\_Health’ from 0.44\% to 0.28\%. 

\subsection*{Entailment and Contradiction Patterns Across Key Attributes}
\begin{figure}[H]
    \centering
    \begin{subfigure}{0.65\textwidth}
        \centering
        \includegraphics[width=\textwidth]{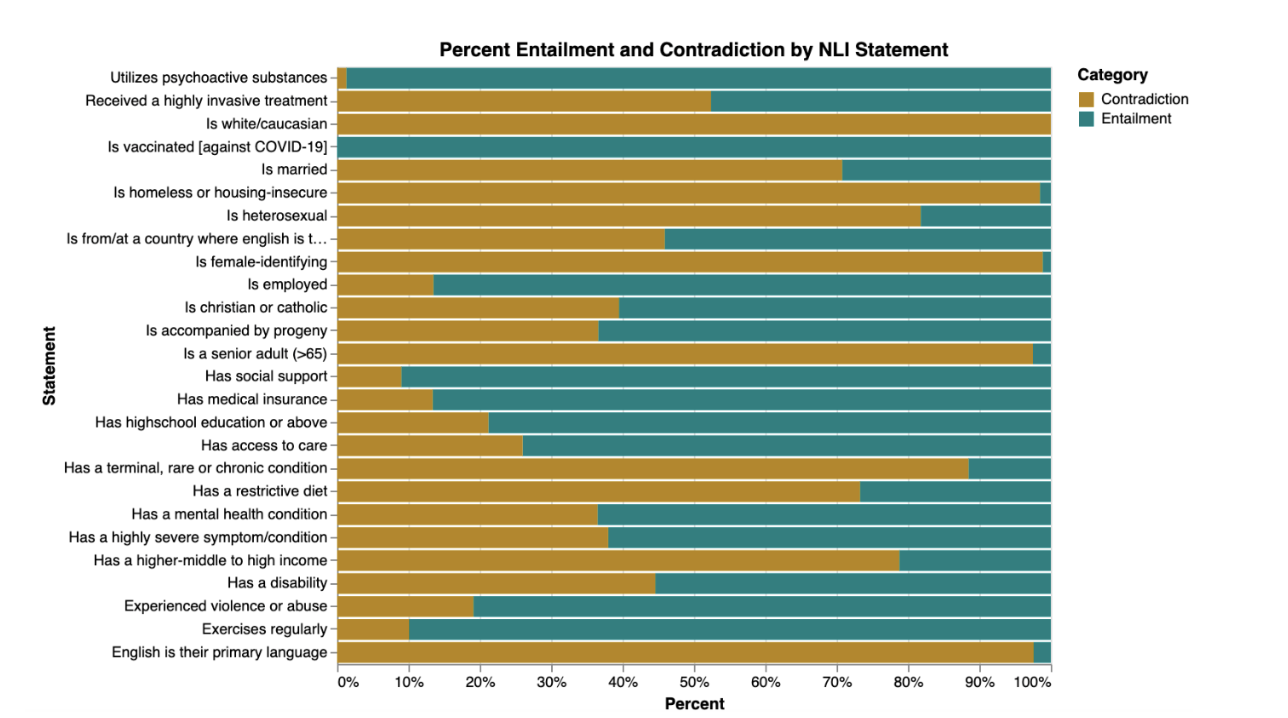}
        \captionsetup{justification=centering}
        \caption{}
    \end{subfigure} 
    \begin{subfigure}{0.65\textwidth}
        \centering
        \includegraphics[width=\textwidth]{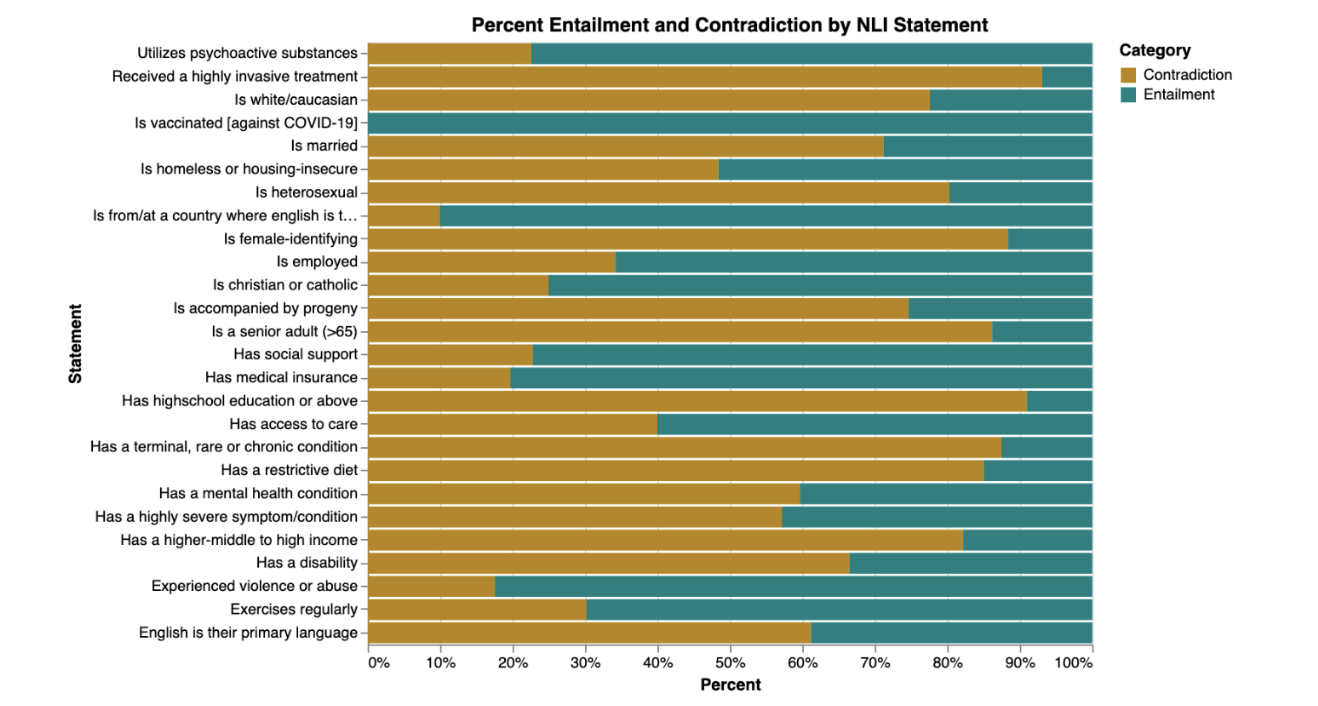}
        \captionsetup{justification=centering}
        \caption{}
    \end{subfigure}
    \caption{(a) Percent Entailment and Contradiction by NLI Statement with RegEx Matching Within Model-Predicted Label Dimensions. (b) Percent entailment and contradiction by NLI statement, with RegEx matching on the full dataset.}  
    \label{fig:4}
\end{figure}

All extracted entities were grouped by their respective entity types and matched with the corresponding statements allowing for a clear comparison of how each entity aligns with or contradicts the defined relationships (Figure 4). The analysis of entailment and contradiction across various entity types reveals several key trends (Fig \ref{fig:4} (a)). Notably, \textit{“Experienced violence or abuse”} and \textit{“Has medical insurance”} are predominantly entailed, at 82.4\% and 80.3\%. Several attributes, including \textit{“Utilizes psychoactive substances,”} \textit{“Has social support,”} \textit{“Exercises regularly,”} \textit{“Is employed,”} and \textit{“Has access to care,”} display moderate levels of entailment while also showing notable contradictions. In contrast, attributes such as \textit{“Is a senior adult,”} \textit{“Is female-identifying,”} \textit{“Is married,”} \textit{“Is heterosexual,”} and \textit{“Has a terminal, rare, or chronic condition”} exhibit high levels of contradiction, with contradiction rates reaching as high as 97.5\%, 98.5\%, 70.8\%, 81.8\%, and 88.5\%. Other attributes, such as \textit{“Is homeless,”} \textit{“Has high school education,”} and \textit{“Is white/Caucasian,”} show more variability in their levels of entailment and contradiction.To further evaluate whether these patterns persist in the full-text context, we removed prior matching restrictions on the extracted entities (Fig \ref{fig:4} (b)). This comparison revealed consistent trends across most entity types, regardless of the applied restrictions.

\section*{Discussion}
Our study demonstrates the potential of LLMs, particularly BERT-based models, in addressing SDOH-specific NER tasks. The use of NLP to gain insights into health-related problems has been an area of research since the 1950s \cite{Wikipedia_contributors2024-sw}, but its large-scale application gained significant momentum during the COVID-19 pandemic. This shift was exemplified by initiatives such as the CORD-19 competition \cite{Allen_Institute_For_AI2022-uz,R_Bhambhoria_L_Feng_D_Sepehr_J_Chen_C_Cowling_S_Kocak_E_Dolatabadi2020-pn}, which coincided with the emergence of transformer-based models \cite{Vaswani2017-qh} and significant advancements in NLP methodologies and applications \cite{Zhao2023-ge}. These developments marked a transformative era in NLP, enabling unprecedented breakthroughs in understanding and analyzing health data \cite{Hu2024-cl,Zhou2024-id}. By proposing a generalizable framework for extracting these attributes, we demonstrate the viability of leveraging LLMs for this purpose. Notably, our findings indicate that BERT-based encoder-only models achieved superior classification performance on the generalization set, underscoring their efficacy in handling complex, domain-specific NER tasks. Previous studies have demonstrated the superior generalization capabilities of BERT models compared to both traditional deep learning and also decoder based transformer models in NER tasks, largely due to their self-attention mechanisms \cite{Mundotiya2024-xd,Obadinma2024-ty}. Our findings align with these benchmarks, providing further evidence of BERT's robustness in handling unseen sentence structures and addressing class sparsity \cite{Devlin2018-st,Sanh2019-me}. In contrast, the comparatively lower performance of the BiomedNLP model is attributed to its pre-training on domain-specific biomedical texts, which limits its flexibility and adaptability \cite{Gu2020-pu}.

The exploratory analysis revealed significant gaps in sociodemographic representation within PCC case reports, with attributes like race, spiritual beliefs, and housing status appearing infrequently. In contrast, age, gender, mental health, and access to care were more frequently discussed. Additionally, the co-occurrence patterns of biomedical and sociodemographic factors often followed a structured narrative, with age and gender commonly contextualizing conditions and symptoms. While case report sections typically contained 18 to 20 non-‘O’ distinct entities dimensions, it was rare for a report to encompass all dimensions; only one of 7,172 reports included predictions for all 26 identified labels. These findings align with prior research highlighting the scarcity of robust sociodemographic representation in clinical PCC datasets \cite{Khunti2024-wf,Prashar2023-fj,Banerjee2024-dd,Hua2024-gd}. Our analysis confirms that academic PCC-related case reports similarly lack diversity across multiple attributes, further underscoring the need for improved documentation and reporting practices to enable more equitable and comprehensive analyses in PCC research. Lastly, an analysis of the most frequent label trigrams reveals that ‘\textit{Condition’} commonly appears in the final position, preceded by entities such as ‘\textit{Age’}, ‘\textit{Gender’}, and ‘\textit{Severity’}. This pattern suggests two key findings: first, academic case reports often adopt a structured approach to representing the patient experience; second, this representation is frequently constrained to a limited subset of sociodemographic factors.

Our key findings highlighted that mentions of attributes such as ‘having access to care’, ‘being vaccinated against COVID-19’, ‘exercising regularly’, ‘having medical insurance’, ‘social support’, and ‘experiencing violence or abuse’ were predominantly in agreement, reflecting positive representation. In contrast, mentions of being ‘female-identifying’, ‘white/Caucasian’, ‘heterosexual’, ‘senior’, or ‘having a terminal, rare, or chronic condition’ were most frequently classified as contradictory to the statement, suggesting potential underreporting or biases. Some of these findings highlight discrepancies with prior research, particularly concerning the underrepresentation or absence of mentions related to being female-identifying, unvaccinated for COVID-19, or a senior adult. Previous studies have identified these groups as potential at-risk populations in the COVID-19 context, with discrimination being significantly associated with prolonged COVID-19 symptoms \cite{Bai2022-wn,Fernandez-de-Las-Penas2022-vt,Thomason2022-hy,Kuy2020-mc,Notarte2022-zq}. Research has also linked female gender and sex to a higher prevalence of PCC \cite{Bai2022-wn,Fernandez-de-Las-Penas2022-vt} and highlighted socioeconomic status as a significant factor in PCC development \cite{Louie2023-rx}. However, these insights should not be interpreted as evidence against previous findings. Instead, they may point to underlying biases in academic research and structural barriers that deter these groups from participating in clinical studies. This aligns with earlier work highlighting selection bias and the underrepresentation of older adults and women in clinical research \cite{Cruz-Jentoft2013-jn,Santos-Casado2019-bq}. Additionally, individuals unvaccinated for COVID-19 often face societal discrimination, which may further impact their access to care and inclusion in research \cite{Shaw2022-ax}. These patterns underscore the need to address systemic biases in data collection and study design to ensure more equitable representation, such as standardizing sociodemographic reporting protocols and promoting the inclusion of underrepresented groups in clinical research.

\subsection*{Limitations and Future work}
While our NLP approach effectively identifies many sociodemographic determinants in a PCC-related corpus, its performance in predicting medical symptoms and conditions remains limited. This challenge arises from the vast diversity of diseases and symptoms, which would require pretraining on a large, domain-specific medical text corpus. Our experiments with BiomedNLP demonstrated low performance in the NER task, however, highlighting the need for better-tailored models. The analysis is inherently influenced by corpus-related biases, including a focus on English-language texts and an overrepresentation of individuals with access to care, in addition to biases associated with GPT. Additionally, while we attempted to comprehensively categorize entailment and contradiction tokens using synonyms, some relevant terms may have been overlooked. Importantly, the distributions highlighted by our analysis should not be interpreted as robust associations but rather as exploratory insights.

To address these limitations, future efforts should focus on enhancing the documentation of underrepresented attributes, such as race, housing status, and sexual orientation, to support equitable AI-driven analyses in PCC research. Future work could involve pretraining a BERT or other transformer models on a larger PCC-specific corpus to improve performance, though this would require significant computational resources. Evaluating additional biomedical models beyond BiomedNLP may also enhance NER outcomes. To better understand interactions between entities, future research could develop an ML-driven relation extraction (RE) pipeline to explore how SDOH factors interrelate. Extending this analysis to other patient populations would provide a broader understanding of sociodemographic mentions across diverse health contexts. Additionally, the insights into PCC-entity representation could inform policy discussions on diversity and inclusion in academic research. Adapting the pipeline to other health domains could also offer a more comprehensive view of sociodemographic reporting in case reports. Finally, given the evolving definition of social determinants of health, all future work must entail an assessment of the relevance and comprehensiveness of the included entity label dimensions.

\section*{Conclusion}
This analysis benchmarks traditional deep learning models and encoder-only transformer models for Named Entity Recognition of sociodemographic entities in academic case reports related to Post COVID-19 Condition. Importantly, it not only provides a model comparison assessment, but provides a comprehensive pipeline equipped with various annotation, augmentation, cleaning and regularization techniques, culminating in a timely model that can be utilized by researchers in the field. Notably, encoder-only transformer models were found to outperform traditional deep learning models on unseen data with distinct sentence structures, and greater class sparsity. Furthermore, the best performing model on the validation data [BERT base uncased] was applied to all 7,172 extracted case reports. Insights were extracted from the annotated data through exploratory techniques and through a rule-based natural language inference (NLI) pipeline leveraging variations generated by GPT-4o for RegEx matching. Ultimately, we conclude that there is scarcity in multiple sociodemographic factors in PCC-related academic case reports, and an imbalanced representation of mentions aligned with groups of interest within dimensions such as gender, insurance status, and age.

\section*{Supporting Information}

\textbf{S1 Appendix A.} Corpus Construction Annotation Details\\
In this section, we provide further details on the entity annotation process for PCC case report corpus.\\

\textbf{S1 A.1} Additional explorations on the corpus with GPT\\

For additional robustness on the alignment with the extracted papers with long COVID, we experimented with prompting OpenAI’s GPT 3.5 Turbo API to classify whether on a subset of 2,000 papers were related to long COVID or not. As we found, however, the model had difficulty classifying the texts in a consistent and robust manner, as the definition of long COVID is highly dependent on the time of the first diagnosis with COVID-19 infection, which is rarely discussed in the case reports.

For the annotation task of subset 1, we explored three different techniques. Initially, we prompted the GPT-3.5 turbo (API) model to classify each token on the case report sections, based on a curated list of labels derived from those indicated by John Snow Labs, literature review and expert advice. Although we found that the model performed well with classification of a few entities, it did not perform well on a multi-class classification task of more than 10 labels.\\

\textbf{S1 A.2} Entity Label Refinement\\
To streamline the entity set, we consolidated closely related labels and removed those deemed non-essential for our use case. This process is outlined in sFigure 1 below.\\

\textbf{S1 A.3} Sample prompt to GPT-4o produce variations per entity type.\\
“For named entity recognition tasks using the CoNLL format, labels are split into B-label and I-label, where B signifies the beginning and I signifies the insight of a label for multi-word entities. Please provide two lists of diverse single-word variations related to this entity type dimension {label}, with this definition {definition}.  One list should include words for B-label and the other list should be for words for I-label. Ensure that there is no overlap between the lists. Here is an example {example}”\\

\textbf{S1 A.4} Dictionary of Variations per Entity Type (Produced by GPT-4o).\\
variations\_rare\_entities=\{'B-Social\_Support':["community", "group", "peer", "mutual", "volunteer", "support", "counseling", "welfare", "social", "charity", "family", "neighborly", "emergency", "relief", "disaster"], 'I-Social\_Support':["service", "therapy", "assistance", "aid", "work", "group", "session", "service", "work", "organization", "support", "help", "shelter", "aid", "response"], 'B-Race\_Ethnicity':["Latino", "Black", "Asian", "White", "Hispanic", "Native", "African", "European", "Arab", "Pacific", "Indigenous", "Mixed", "Jewish", "Indian", "Korean"], 'I-Race\_Ethnicity':["American", "Canadian", "British", "Australian", "Islander", "Peoples", "Race", "Community", "Descent", "Ethnicity", "Heritage", "Culture", "Group", "Population", "Minority"],'B-Exercise':["running", "swimming", "cycling", "weightlifting", "yoga", "aerobics", "pilates", "hiking", "dancing", "jogging", "boxing", "rowing", "climbing", "jumping", "skipping"],'I-Exercise':["routine", "session", "workout", "practice", "training", "regimen", "activity", "drill", "movement", "exercise", "stretch", "program", "set", "circuit", "plan"], 'B-Education':["school", "college", "university", "academy", "institute", "kindergarten", "preschool", "highschool", "middle", "elementary", "primary", "secondary", "vocational", "training", "seminar"],  'I-Education':["degree", "diploma", "course", "program", "class", "lesson", "certificate", "studies", "education", "learning", "major", "specialization", "qualification", "curriculum", "instruction"], 'B-Geographic\_Entity':["Paris", "Canada", "Tokyo", "Brazil", "Berlin", "India", "London", "China", "Sydney", "Mexico", "Rome", "Russia", "Chicago", "Egypt", "Madrid"], 'I-Geographic\_Entity':["province", "capital", "metropolis", "country", "district", "suburb", "state", "township", "territory", "neighborhood", "island", "county", "territory", "zone", "area"], 'B-Substance':["alcohol", "cigarette", "marijuana", "cocaine", "heroin", "methamphetamine", "ecstasy", "LSD", "psilocybin", "ketamine", "opioid", "amphetamine", "tobacco", "caffeine", "nicotine"], 'I-Substance':["use", "consumption", "dependence", "addiction", "misuse", "habit", "intoxication", "ingestion", "administration", "dose", "exposure", "indulgence", "practice", "dependence"],'B-Severity':["severe", "critical", "extreme", "acute", "intense", "serious", "grave", "significant", "major", "urgent", "critical", "drastic", "dire", "severe", "crucial"], 'I-Severity':["condition", "case", "situation", "stage", "level", "instance", "circumstance", "state", "point", "stage", "level", "degree", "severity", "intensity", "extent"], 'B-Marital\_Status':["single", "married", "divorced", "widowed", "separated", "engaged", "partnered", "unmarried", "committed", "relationship", "civil", "union", "spouse", "marriage"], 'I-Marital\_Status':["status", "relationship"], 'B-Disability':["physical", "intellectual", "developmental", "mental", "emotional", "sensory", "cognitive", "psychological", "chronic", "visible", "invisible", "permanent", "temporary", "disability", "impairment"], 'I-Disability':["condition", "disorder", "issue", "challenge", "disability", "impairment", "problem", "limitation", "affliction", "ailment", "handicap", "constraint", "barrier", "hurdle", "obstacle"], 'B-Housing':["homeless", "renter", "homeowner", "tenant", "landlord", "lodger", "resident", "dweller", "occupant", "house", "apartment", "condominium", "townhouse", "villa", "shelter"], 'I-Housing':["living", "situation", "arrangement", "status", "condition", "accommodation", "housing", "unit", "space", "place", "option", "residence", "dwelling", "premises", "property"], 'B-Diet':["vegan", "vegetarian", "pescatarian", "carnivore", "omnivore", "keto", "paleo", "gluten-free", "low-carb", "low-fat", "high-protein", "balanced", "raw", "plant-based", "whole-food"], 'I-Diet':["diet", "plan", "regimen", "nutrition", "eating", "routine", "lifestyle", "habit", "practice", "approach", "program", "style", "protocol", "method", "choice"], 'B-Income':["income", "salary", "wage", "earnings", "pay", "compensation", "revenue", "stipend", "remuneration", "profit", "compensation", "recompense", "gain", "return", "receipt"], 'I-Income':["level", "bracket", "range", "category", "tier", "class", "scale", "segment", "division", "section", "grade", "classification", "band", "division", "stratum"], 'B-Language':["english", "spanish", "french", "mandarin", "arabic", "hindi", "bengali", "portuguese", "russian", "japanese", "german", "korean", "italian", "turkish", "vietnamese"], 'I-Language':["language", "dialect", "vernacular", "tongue", "speech", "communication", "idiom", "lingo", "jargon", "vocabulary", "lexicon", "phraseology", "linguistics", "grammar", "accent"], 'B-Insurance\_Status':["insured", "uninsured", "covered", "protected", "policyholder", "enrolled", "subscriber", "beneficiary", "underwritten", "guaranteed", "assured", "insurer", "coverage", "insurance", "plan"], 'I-Insurance\_Status':["status", "holder", "member", "recipient", "participating", "enrollee", "policy", "participant", "benefit", "coverage", "plan", "option", "program", "scheme", "arrangement"], 'B-Violence\_Or\_Abuse':["violence", "abuse", "assault", "battery", "harassment", "bullying", "intimidation", "coercion", "exploitation", "neglect", "trauma", "maltreatment", "domestic", "partner", "stalking"],  'I-Violence\_Or\_Abuse':["incident", "case", "act", "behavior", "occurrence", "confrontation", "abuse", "encounter", "occurrence"], 'B-Family\_Member':["father", "mother", "brother", "sister", "grandfather", "grandmother", "uncle", "aunt", "cousin", "son", "daughter", "nephew", "niece", "husband", "wife"], 'I-Family\_Member':["parent", "sibling", "grandparent", "relative", "family", "member", "child", "offspring", "kin", "spouse", "partner", "relation", "in-law", "step-", "adoptive"],  'B-Sexual\_Orientation':["gay", "lesbian", "bisexual", "heterosexual", "queer", "pansexual", "asexual", "demisexual", "homosexual", "biromantic", "heteroromantic", "sapiosexual", "aromantic", "bicurious", "questioning"], 'I-Sexual\_Orientation':["orientation", "identity", "preference", "inclination", "attraction", "orientation", "orientation", "identity", "orientation", "identity", "preference", "orientation", "identity", "orientation", "identity"], 'B-Gender':["man", "woman", "male", "female", "boy", "girl", "transgender", "nonbinary", "agender", "bigender", "cisgender", "genderfluid", "two-spirit"], 'I-Gender':["identity", "identifying"], 'B-Spiritual\_Beliefs':["christian", "islamic", "buddhist", "hindu", "jewish", "sikh", "taoist", "shinto", "bahai", "jain", "pagan", "wiccan", "new\_age", "atheist", "agnostic"], 'I-Spiritual\_Beliefs':["belief", "faith", "religion", "spirituality", "doctrine", "philosophy", "creed", "teaching", "tenet", "principle", "ideology", "conviction", "dogma", "mythology", "doctrine"],'B-Age' : ["young", "adult", "elderly", "child", "teenage", "middle-aged", "senior", "adolescent", "infant", "toddler", "youthful", "mature", "50-year-old", "30-year-old", "20-year-old", "10-year-old", "newborn"],'I-Age': ["years", "old", "age"],'B-Employment': ["employed", "unemployed", "worker", "employee", "jobholder", "laborer", "professional", "executive", "manager", "entrepreneur", "freelancer", "contractor", "temp", "part-time", "full-time", "doctor", "teacher", "engineer", "artist", "chef"], 'I-Employment': ["position", "role", "occupation", "job", "career", "work"], 'B-Vaccine': ["pfizer", "moderna", "astrazeneca", "johnson\_and\_johnson", "sinovac", "sinopharm", "sputnik", "vaccinated", "unvaccinated", "immunized"], 'I-Vaccine': ["vaccination", "immunization", "dose", "inoculation","vaccine"], 'B-Mental\_Health': ["depression", "anxiety", "bipolar", "schizophrenia", "ptsd", "OCD", "psychosis", "happiness", "self-care", "mindfulness", "psychiatric", "therapy"], 'I-Mental\_Health': ["state"], 'B-Treatment':["chemotherapy", "radiation", "dialysis", "transplant", "vaccination", "immunotherapy", "antibiotics", "antivirals", "physiotherapy", "chiropractic", "acupuncture", "orthopedic", "dermatological", "cardiac", "oncological", "surgery", "rehabilitation", "intervention", "treatment", "holistic", "palliative", "homeopathy"], 'I-Treatment': ["plan", "session", "regimen", "protocol", "procedure", "approach", "method", "intervention", "strategy", "technique", "therapy"], 'B-Access\_To\_Care': ["availability", "proximity", "affordability", "coverage", "accessibility", "transportation", "providers", "appointments", "clinics", "hospitals", "facilities", "ICU", "telehealth"],'I-Access\_To\_Care' :[ "barrier", "waitlist", "delay", "shortage", "gap", "disparity", "limitation", "restriction", "obstacle", "constraint"],"B-Condition":["asthma", "diabetes", "hypertension", "arthritis", "cancer", "stroke", "epilepsy", "dementia", "migraine", "cardiomyopathy", "anemia", "osteoporosis", "glaucoma", "parkinson"],"I-Condition":["diagnosis", "stage", "symptom", "manifestation", "prognosis", "complication", "episode", "onset", "recurrence", "chronic", "persistent", "latent", "remission"]\}\\

\textbf{S1 A.5} Sample prompt to generate sentence structures for augmentation of the development set.\\

“For a named entity recognition task using the CoNLL format, the following labels are utilized \{set\_of\_non\_o\_labels\}. Here are the definitions of each label dimension \{definitions\}. 

Please generate three distinct sentence structures using all the provided labels once (in total). They should follow these conditions:

1) Labels being related to same entity dimension must follow each other in the order B-label I-label

2) Each sentence structure must have a different context and label order than the others

3) All spaces where the labels would occur are maintained with the label names

4) All sentences should include a minimal number of words outside of the labels, while maintaining contextual integrity

Here is an example: \{example\}”

\textbf{S1 A.6} Sample prompt to generate sentence structures for augmentation of the generalization set.

“For a named entity recognition task using the CoNLL format, the following labels are utilized \{set\_of\_non\_o\_labels\}. Here are the definitions of each label dimension \{definitions\}. 

Please generate a set of three distinct sentence structures using all the provided labels once (in total). They should follow these conditions:

1) Labels being related to same entity dimension must follow each other in the order B-label I-label

2) Each sentence structure must have a different context and label order than the others

3) All spaces where the labels would occur are maintained with the label names

4) All sentence structures should have a distinct label order and a larger number of words than these sentence structures: \{previous\_sentence\_structures\}

Here is an example: \{example\}”\\

\textbf{S1 Appendix B.} SDOH Extraction and Analysis Pipeline Details.\\
In this section we provide additional information on the NER and NLI pipelines.\\

\textbf{S1 B.1} Sample Prompts to GPT-4o for entailment and contradiction set generation\\

“Please provide a list of all single words related to this \{entity\_label\_class\} that make the following statement true: \{Statement\}”\\

“Now please provide a list of all single words related to this \{entity\_label\_class\} that are contradictory the following statement: \{Statement\}”\\

\textbf{S1 B.2} NLI entailment and contradiction sets by entity type (generated by GPT-4o)\\

"Access To Care":{"entailment:[
    "clinic",
    "hospital",
    "doctor",
    "nurse",
    "appointment",
    "treatment",
    "medication",
    "pharmacy",
    "insurance",
    "provider",
    "specialist",
    "diagnosis",
    "screening",
    "therapy",
    "consultation",
    "referral",
    "prescription",
    "coverage",
    "caregiver",
    "facility",
    "emergency",
    "service",
    "healthcare",
    "benefit",
    "ICU",
    "intake",
    "admitted",
    "scheduled",
    "enter"
], 'contradiction':[
    "barrier",
    "delay",
    "unavailable",
    "uninsured",
    "inaccessible",
    "cost",
    "expense",
    "inequality",
    "disparity",
    "distance",
    "waitlist",
    "overcrowding",
    "outreach",
    "underserved",
    "limited",
    "restriction",
    "shortage",
    "neglect",
    "exclusion",
    "inequity"]},'Age':{'entailment':['65',
 '66',
 '67',
 '68',
 '69',
 '70',
 '71',
 '72',
 '73',
 '74',
 '75',
 '76',
 '77',
 '78',
 '79',
 '80',
 '81',
 '82',
 '83',
 '84',
 '85',
 '86',
 '87',
 '88',
 '89',
 '90',
 '91',
 '92',
 '93',
 '94',
 '95',
 '96',
 '97',
 '98',
 '99',
 '100',
 '101',
 '102',
 '103',
 '104',
 '105',
 '106',
 '107',
 '108',
 '109',
 '110',
 '111',
 '112',
 '113',
 '114',
 '115',
 '116',
 '117',
 '118',
 '119',
 'senior',
 'elderly',
 'aged',
 'senior citizen',
 'retiree',
 'geriatric',
 'old',
 'mature',
 'veteran',
 'ancient',
 'pensioner',
 'octogenarian',
 'nonagenarian',
 'centenarian'], 'contradiction':['1',
 '2',
 '3',
 '4',
 '5',
 '6',
 '7',
 '8',
 '9',
 '10',
 '11',
 '12',
 '13',
 '14',
 '15',
 '16',
 '17',
 '18',
 '19',
 '20',
 '21',
 '22',
 '23',
 '24',
 '25',
 '26',
 '27',
 '28',
 '29',
 '30',
 '31',
 '32',
 '33',
 '34',
 '35',
 '36',
 '37',
 '38',
 '39',
 '40',
 '41',
 '42',
 '43',
 '44',
 '45',
 '46',
 '47',
 '48',
 '49',
 '50',
 '51',
 '52',
 '53',
 '54',
 '55',
 '56',
 '57',
 '58',
 '59',
 '60',
 '61',
 '62',
 '63',
 'young',
 'youthful',
 'adolescent',
 'teenager',
 'juvenile',
 'minor',
 'child',
 'preteen',
 'teen',
 'toddler',
 'infant',
 'preschooler',
 'youngster',
 'adult',
 'youth',
 'preadult',
 'earlyadult']},
 'Condition':{'entailment':[
    "cancer",
    "diabetes",
    "fibrosis",
    "Huntington",
    "Parkinson",
    "ALS",
    "dementia",
    "epilepsy",
    "HIV",
    "AIDS",
    "lupus",
    "sarcoma",
    "leukemia",
    "lymphoma",
    "hemophilia",
    "celiac",
    "Crohn",
    "colitis",
    "sclerosis",
    "malaria",
    "tuberculosis",
    "Alzheimer",
    "thalassemia",
    "cystinosis",
    "marfan",
    "sickle",
    "lyme",
    "endometriosis",
    "myeloma",
    "amyloidosis",
    "neurofibromatosis",
    "phenylketonuria",
    "porphyria",
    "retinoblastoma",
    "scleroderma",
    "spina",
    "muscular",
    "dystrophy",
    "hypertension",
    "cardiomyopathy",
    "hepatitis",
    "cirrhosis",
    "nephropathy",
    "uremia",
    "glioblastoma"
], 'contradiction':[
    "headache",
    "fever",
    "cough",
    "nausea",
    "fatigue",
    "dizziness",
    "sore",
    "itching",
    "rash",
    "cold",
    "flu",
    "allergy",
    "sprain",
    "bruise",
    "infection",
    "constipation",
    "diarrhea",
    "vomiting",
    "stomachache",
    "indigestion",
    "burn",
    "cut",
    "blister",
    "acne",
    "cramp",
    "fracture",
    "swelling",
    "nosebleed",
    "sneezing",
    "hiccup",
    "thirst",
    "dehydration",
    "insomnia",
    "anxiety",
    "stress",
    "tremor",
    "irritation",
    "dryness",
    "fainting",
    "bloating",
    "heartburn",
    "itchiness",
    "runny",
    "congestion",
    "sensitivity",
    "numbness"
]},
 'Diet':{'entailment':[
    "vegan",
    "vegetarian",
    "paleo",
    "keto",
    "gluten-free",
    "dairy-free",
    "halal",
    "kosher",
    "pescatarian",
    "low-carb",
    "low-fat",
    "low-sodium",
    "mediterranean",
    "carnivore",
    "macrobiotic",
    "whole30",
    "alkaline",
    "raw",
    "frugivore",
    "fasting"
], 'contradiction':[
    "omnivore",
    "balanced",
    "flexitarian",
    "standard",
    "mixed",
    "unrestricted",
    "traditional",
    "casual",
    "regular",
    "moderate"
]},
 'Disability':{'entailment':[
    "disabled",
    "person-with-disability",
    "individual-with-disability",
    "differently-abled",
    "challenged",
    "handicapped",
    "impaired",
    "physically-disabled",
    "mentally-disabled",
    "visual-impairment",
    "hearing-impairment",
    "deaf",
    "blind",
    "partially-sighted",
    "low-vision",
    "mobility-impairment",
    "wheelchair-user",
    "spinal-cord-injury",
    "amputee",
    "cerebral-palsy",
    "multiple-sclerosis",
    "muscular-dystrophy",
    "autism-spectrum-disorder",
    "ADHD",
    "attention-deficit-disorder",
    "intellectual-disability",
    "developmental-disability",
    "epilepsy",
    "chronic-pain",
    "chronic-fatigue-syndrome",
    "fibromyalgia",
    "mental-health-condition",
    "bipolar-disorder",
    "schizophrenia",
    "depression",
    "anxiety-disorder",
    "PTSD",
    "post-traumatic-stress-disorder"
], "contradiction":[
    "able-bodied",
    "able",
    "without-disability",
    "fully-abled",
    "healthy",
    "functioning-normally",
    "typically-abled",
    "normal",
    "intact",
    "capable",
    "unrestricted",
    "unaffected",
    "well",
    "fit",
    "active",
    "independent",
    "non-impaired",
    "unrestricted-functioning",
    "without-condition",
    "optimal-health",
    "normal-functioning"]},
 "Education":{"entailment":[
    "alumni",
    "assignment",
    "bachelor",
    "capstone",
    "credit",
    "diploma",
    "dissertation",
    "enrollment",
    "graduation",
    "honors",
    "internship",
    "major",
    "master's",
    "prerequisite",
    "research",
    "semester",
    "sophomore",
    "thesis",
    "transcript",
    "undergraduate",
    "university",
    "vocational",
    "doctor",
    "engineer",
    "nurse",
    "specialist",
    "GED",
    "lawyer"
], "contradiction":[
    "basic",
    "classroom",
    "concept",
    "elementary",
    "exercise",
    "homework",
    "schoolwork",
    "young",
    "child",
    "infant",
    "unskilled",
    "illiterate",
    "middleschool",
    "kindergarten",
    "baby",
    "childcare",
    "play",
    "uneducated",
    "alphabet",
    "book",
    "class",
    "coloring",
    "daily",
    "drawing",
    "interactive",
    "literacy",
    "map",
    "quiz"
]},
 "Employment":{"entailment":[
    "accountant",
    "application",
    "benefits",
    "boss",
    "career",
    "contract",
    "coworker",
    "daycare",
    "employee",
    "employer",
    "engineer",
    "hire",
    "intern",
    "job",
    "manager",
    "nurse",
    "office",
    "payroll",
    "position",
    "promotion",
    "recruiter",
    "resume",
    "salary",
    "secretary",
    "staff",
    "supervisor",
    "task",
    "team",
    "technician",
    "trainer",
    "vacation",
    "wage",
    "doctor",
    "nurse"
], "contradiction":[
    "absence",
    "benefits",
    "crisis",
    "downtime",
    "furlough",
    "jobless",
    "layoff",
    "loss",
    "out-of-work",
    "recession",
    "redundancy",
    "resignation",
    "severance",
    "situation",
    "suspension",
    "termination",
    "unemployed",
    "vacant",
    "welfare",
    "without",
    "retraining",
    "searching",
    "displacement"
]},
 "Exercise":{"entailment":[
    "Cardio",
    "Jogging",
    "Cycling",
    "Swimming",
    "Rowing",
    "CrossFit",
    "Lifting",
    "Pilates",
    "Yoga",
    "Running",
    "Drills",
    "Training",
    "Conditioning",
    "Workouts",
    "Mobility",
    "Aerobics"
], "contradiction":[
    "Sedentary",
    "Idle",
    "Inactive",
    "Dormant",
    "Lazy",
    "Unfit",
    "Neglect",
    "Lethargic",
    "Sluggish",
    "Unmotivated",
    "Resting",
    "Indolent",
    "Procrastination",
    "Apathetic",
    "Static",
    "Stagnant",
    "Leisurely",
    "Avoidance",
    "Neglectful",
    "Out-of-shape"
]},
 "Family Member":{"entailment":[
    "child",
    "son",
    "daughter"
], "contradiction":[
    "cousin",
    "sister",
    "brother",
    "sibling",
    "grandma",
    "grandpa"
]},
 "Gender":{"entailment":[
    "Woman",
    "Female",
    "Girl",
    "Lady",
    "Wife",
    "Daughter",
    "Gal",
    "Miss",
    "Sister",
    "Queen",
    "Matriarch",
    "Mother",
    "Diva",
    "Chick",
    "Ma'am",
    "she",
    "girl",
    "grandmother",
    "feminine",
    "sister"
], "contradiction":[
    "Man",
    "Male",
    "Boy",
    "Gentleman",
    "Husband",
    "Son",
    "Dude",
    "Guy",
    "Brother",
    "Patriarch",
    "Father",
    "Macho",
    "he",
    "Nonbinary",
    "Genderqueer",
    "Genderfluid",
    "Agender",
    "Bigender", 
    "Demiboy"
]},
 "Geographic Entity":{"entailment":[
    "England",
    "Scotland",
    "Wales",
    "Ireland",
    "USA",
    "Canada",
    "Australia",
    "New Zealand",
    "London",
    "Sydney",
    "Toronto",
    "Dublin",
    "Edinburgh",
    "Melbourne",
    "Vancouver",
    "Glasgow",
    "Auckland",
    "Houston",
    "Chicago",
    "San Francisco",
    "New York",
    "Los Angeles",
    "Boston",
    "Seattle",
    "Brisbane",
    "Perth",
    "Calgary",
    "Ottawa",
    "Manchester",
    "Belfast",
    "Wellington",
    "Adelaide",
    "Kansas City",
    "Philadelphia",
    "Atlanta",
    "San Diego",
    "Minneapolis",
    "Baltimore",
    "United States",
    "New York",
    "State",
    "Province"], "contradiction":[
    "France",
    "Germany",
    "Spain",
    "Italy",
    "China",
    "Japan",
    "Russia",
    "Brazil",
    "South Korea",
    "Mexico",
    "India",
    "Argentina",
    "Turkey",
    "Saudi Arabia",
    "Thailand",
    "Sweden",
    "Norway",
    "Denmark",
    "Greece",
    "Netherlands",
    "Portugal",
    "Poland",
    "Czechia", 
    "Hungary",
    "Vietnam",
    "Malaysia",
    "Indonesia",
    "Egypt",
    "Chile",
    "Colombia",
    "Iran",
    "Israel",
    "Pakistan",
    "Bangladesh",
    "Ukraine",
    "Peru",
    "Romania",
    "Jordan",
    "Philippines",
    "Sri Lanka",
    "Nepal",
    "UAE"  
]},
 "Housing":{"entailment":[
    "Homeless",
    "Displaced",
    "Vagrant",
    "Transient",
    "Squatter",
    "Houseless",
    "Destitute",
    "Unhoused",
    "Wanderer",
    "Nomadic",
    "Refugee",
    "Drifter",
    "Impoverished",
    "Indigent",
    "Evicted",
    "Exiled",
    "Unsheltered",
    "Rootless",
    "Vagabond",
    "Outcast",
    "Shelter",
    "Temporary"
], "contradiction":[
    "Housed",
    "Settled",
    "Lodged",
    "Dweller",
    "Tenant",
    "Homeowner",
    "Proprietor",
    "Lodger",
    "Boarder",
    "Leaseholder",
    "Renter",
    "Landlord",
    "Landlady",
    "Householder",
    "Proprietor",
    "house",
    "apartment",
    "condo",
    "villa",
    "cottage"
]},
 "Income":{"entailment":[
    "high-income",
    "Wealthy",
    "Affluent",
    "Prosperous",
    "Rich",
    "Well-off",
    "Loaded",
    "Flourishing",
    "Opulent",
    "Lavish",
    "Thriving",
    "Comfortable",
    "Elite",
    "Privileged",
    "Millionaire",
    "Billionaire",
    "Affluence",
    "Tycoon",
    "Magnate",
    "Aristocrat",
    "Plutocrat"
], "contradiction":[
    "Poor",
    "Impoverished",
    "Needy",
    "Struggling",
    "Destitute",
    "Low-income",
    "Underprivileged",
    "Broke",
    "Indigent",
    "Disadvantaged",
    "Penniless",
    "Hard-up",
    "Hand-to-mouth",
    "Economizing",
    "Frugal",
    "Working-class",
    "Modest",
    "Income-restricted",
    "Underpaid",
    "Subsidized"
]},
 "Insurance Status":{"entailment":[
    "insured",
    "covered",
    "protected",
    "policyholder",
    "subscriber",
    "beneficiary",
    "enrolled",
    "planholder",
    "participant",
    "member",
    "certificate-holder",
    "dependent",
    "claimant",
    "premium-payer",
    "co-insured",
    "beneficiary-owner",
    "enrollee",
    "recipient",
    "policy-beneficiary",
    "co-pay-holder"
], "contradiction":[
    "uninsured",
    "unprotected",
    "uncovered",
    "self-pay",
    "out-of-pocket",
    "non-covered",
    "non-insured",
    "underinsured",
    "vulnerable",
    "unsubscribed",
    "unregistered",
    "non-enrolled",
    "policy-lapsed",
    "unaffiliated",
    "non-member",
    "not-covered",
    "excluded",
    "no-coverage"
]},
    "Language":{"entailment":[
    "anglophone",
    "english-speaking",
    "english-talking",
    "english-fluent",
    "english-proficient",
    "english-literate",
    "english-user",
    "english-native",
    "english-conversant",
    "english-articulate",
    "english-communicator",
    "english",
    "biligual"
], "contradiction":[
    "spanish",
    "french",
    "german",
    "mandarin",
    "arabic",
    "russian",
    "hindi",
    "japanese",
    "korean",
    "italian",
    "portuguese",
    "swahili",
    "urdu",
    "bengali",
    "vietnamese",
    "turkish",
    "thai",
    "persian",
    "polish",
    "dutch",
    "greek",
    "hebrew",
    "malay",
    "tagalog",
    "indonesian",
    "tamil",
    "telugu",
    "swedish",
    "norwegian",
    "finnish",
    "danish",
    "punjabi",
    "gujarati",
    "cantonese",
    "ukrainian",
    "hungarian",
    "romanian",
    "czech",
    "slovak",
    "serbian",
    "croatian",
    "bosnian",
    "bulgarian",
    "albanian",
    "georgian",
    "armenian",
    "kurdish",
    "pashto",
    "somali",
    "zulu",
    "amharic",
    "yiddish"
]},
 "Marital Status":{"entailment":[
    "married",
    "wed",
    "spouse",
    "husband",
    "wife",
    "partnered",
    "wedded",
    "hitched",
    "betrothed",
    "matrimonial",
    "nuptial",
    "coupled",
    "joined",
    "in-union",
    "espoused",
    "conjugally-bound",
    "life-partner",
    "in-wedlock",
    "legally-bound",
    "spousal"
], "contradiction":[
    "unmarried",
    "single",
    "bachelor",
    "bachelorette",
    "divorced",
    "widowed",
    "separated",
    "not-in-union",
    "unwed",
    "not-married",
    "solo",
    "independent",
    "available",
    "unpartnered",
    "single-status",
    "unconjugated",
    "not-attached",
    "non-marital",
    "single-person",
    "uncommitted"
]},
 "Mental Health":{"entailment":[
    "anxiety-disorder",
    "depression",
    "bipolar-disorder",
    "schizophrenia",
    "obsessive-compulsive-disorder",
    "post-traumatic-stress-disorder",
    "attention-deficit-hyperactivity-disorder",
    "borderline-personality-disorder",
    "eating-disorder",
    "panic-disorder",
    "social-anxiety-disorder",
    "generalized-anxiety-disorder",
    "schizoaffective-disorder",
    "autism-spectrum-disorder",
    "dysthymia",
    "seasonal-affective-disorder",
    "psychotic-disorder",
    "somatic-symptom-disorder",
    "dissociative-identity-disorder",
    "paranoid-personality-disorder",
    "narcissistic-personality-disorder",
    "avoidant-personality-disorder",
    "dependent-personality-disorder",
    "histrionic-personality-disorder",
    "substance-use-disorder",
    "sleep-disorder",
    "bipolar-II-disorder",
    "premenstrual-dysphoric-disorder",
    "trichotillomania",
    "hoarding-disorder",
    "impulse-control-disorder",
    "delusional-disorder",
    "antisocial-personality-disorder",
    "psychotic-break",
    "complex-post-traumatic-stress-disorder",
    "neurocognitive-disorder",
    "paranoid-schizophrenia",
    "schizophreniform-disorder",
    "reactive-attachment-disorder",
    "selective-mutism",
    "gender-dysphoria"
], "contradiction":[
    "emotional-regulation",
    "self-care",
    "mindfulness",
    "emotional-health",
    "self-esteem",
    "mental-fitness",
    "meditation",
    "self-awareness",
    "adjustment",
    "relaxation",
    "emotional-support",
    "mental-clarity",
    "behavioral-health",
    "personal-growth",
    "emotional-resilience",
    "stress-management",
    "mental-state",
    "emotional-wellness",
    "life-skills",
    "self-help",
    "emotional-intelligence",
    "wellness",
    "positive-mental-health"
]},
 "Race Ethnicity":{"entailment":[
    "white",
    "caucasian",
    "european",
    "anglo",
    "non-hispanic-white",
    "of-european-descent",
    "light-skinned",
    "fair-skinned",
    "euro-american",
    "euro-descendant",
    "western-european",
    "ethnically-white",
    "white-american",
    "white-european",
    "caucasoid",
    "british",
    "irish",
    "german",
    "french",
    "italian",
    "spanish",
    "portuguese",
    "scandinavian",
    "dutch",
    "belgian",
    "austrian",
    "swiss",
    "polish",
    "hungarian",
    "romanian",
    "ukrainian",
    "greek",
    "bulgarian",
    "serbian",
    "croatian",
    "slovak",
    "czech",
    "slovenian",
    "estonian",
    "latvian",
    "lithuanian",
    "russian",
    "georgian",
    "armenian"
], "contradiction":[
    "black",
    "african-american",
    "african",
    "hispanic",
    "latino",
    "latina",
    "asian",
    "pacific-islander",
    "native-american",
    "indigenous",
    "middle-eastern",
    "arab",
    "south-asian",
    "east-asian",
    "southeast-asian",
    "native-hawaiian",
    "american-indian",
    "alaskan-native",
    "mixed-race",
    "biracial",
    "multiracial",
    "indian",
    "pakistani",
    "bangladeshi",
    "nepalese",
    "tibetan",
    "mongolian",
    "samoan",
    "tongan",
    "fijian",
    "maori",
    "aboriginal",
    "torres-strait-islander",
    "latinx",
    "nigerian",
    "kenyan",
    "ghanaian",
    "south-african",
    "jamaican",
    "brazilian",
    "colombian",
    "argentinian",
    "peruvian",
    "chilean",
    "japanese",
    "south-korean",
    "vietnamese",
    "filipino",
    "myanmar",
    "laotian",
    "cambodian",
    "fijian",
    "papuan",
    "aboriginal-australian",
    "torres-strait-islander",
    "saudi-arabian",
    "iranian",
    "iraqi",
    "jordanian",
    "egyptian",
    "lebanese",
    "syrian",
    'chi',
    'chin',
    'chun',
    'chung',
     'hong',
    'jia',
    'joo',
    'ko',
    'kong',
    'ku',
     'wang',
     'wet',
 'wong',
 'wu',
 'xu',
 'yue'
]},
 "Severity":{"entailment":[
    "severe",
    "acute",
    "critical",
    "life-threatening",
    "grave",
    "serious",
    "debilitating",
    "intense",
    "extreme",
    "advanced",
    "terminal",
    "dangerous",
    "worsening",
    "incapacitating",
    "drastic",
    "progressive",
    "persistent",
    "unmanageable",
    "high-risk",
    "fatal",
    "irreversible",
    "end-stage",
    "excruciating",
    "complicated",
    "intractable",
    "malignant",
    "emergency",
    "unstable",
    "alarming",
    "deteriorating",
    "profound"
], "contradiction":[
    "mild",
    "minor",
    "moderate",
    "manageable",
    "low-risk",
    "non-serious",
    "stable",
    "temporary",
    "recoverable",
    "benign",
    "treatable",
    "contained",
    "localized",
    "slight",
    "limited",
    "minimal",
    "controlled",
    "subtle",
    "non-life-threatening",
    "infrequent",
    "improving",
    "marginal",
    "relieving",
    "reversible",
    "light"
]},
 "Sexual Orientation":{"entailment":[
    "heterosexual",
    "straight",
    "opposite-sex-attracted",
    "hetero",
    "male-female",
    "heteronormative",
    "binary-attracted",
    "heterosexual-couple",
    "man-woman-relationship",
    "traditional-relationship",
    "hetero-cisgender"
], "contradiction":[
    "homosexual",
    "gay",
    "lesbian",
    "bisexual",
    "pansexual",
    "queer",
    "asexual",
    "demisexual",
    "polysexual",
    "omnisexual",
    "non-binary-attracted",
    "same-sex-attracted",
    "genderqueer-attracted",
    "fluid-sexuality",
    "two-spirit",
    "androgynous-attracted",
    "sapphic",
    "achillean"
]},
 "Social Support":{"entailment":[
    "support",
    "help",
    "assistance",
    "guidance",
    "community",
    "network",
    "companionship",
    "friendship",
    "family-support",
    "emotional-support",
    "counsel",
    "peer-support",
    "mentorship",
    "encouragement",
    "aid",
    "backup",
    "care",
    "nurturing",
    "reassurance",
    "solidarity",
    "comfort",
    "advocacy",
    "protection",
    "cooperation",
    "teamwork",
    "social-network",
    "collaboration",
    "partnership",
    "shared-responsibility",
    "associates",
    "alliance"
], "contradiction":[
    "isolation",
    "loneliness",
    "neglect",
    "abandonment",
    "alienation",
    "disconnection",
    "estrangement",
    "solitude",
    "detachment",
    "lack-of-support",
    "unsupported",
    "unassisted",
    "neglected",
    "forsaken",
    "social-exclusion",
    "marginalization",
    "rejection",
    "disassociation",
    "disengagement",
    "seclusion",
    "friendless",
    "abandoned",
    "unhelped",
    "solitary",
    "outcast",
    "unprotected",
    "uncared-for",
    "disempowered",
    "isolated",
    "unsupported"
]},
 "Spiritual Beliefs":{"entailment":[
    "christian",
    "catholic",
    "christianity",
    "roman-catholic",
    "baptist",
    "protestant",
    "evangelical",
    "methodist",
    "presbyterian",
    "orthodox-christian",
    "lutheran",
    "anglican",
    "episcopalian",
    "born-again",
    "follower-of-christ",
    "believer",
    "churchgoer",
    "disciple",
    "catholicism",
    "christ-follower",
    "christendom",
    "clergy",
    "parishioner",
    "faithful",
    "apostolic",
    "orthodox",
    "jesus-follower",
    "christian-believer"
], "contradiction":[
    "islam",
    "muslim",
    "judaism",
    "jewish",
    "hindu",
    "hinduism",
    "buddhism",
    "buddhist",
    "sikhism",
    "sikh",
    "jainism",
    "jain",
    "taoism",
    "taoist",
    "shinto",
    "shintoism",
    "confucianism",
    "zoroastrianism",
    "zoroastrian",
    "pagan",
    "wiccan",
    "animism",
    "bahá'í",
    "rastafarian",
    "indigenous-religions",
    "shamanism",
    "druze",
    "bahá'í-faith",
    "neo-pagan",
    "voodoo",
    "santeria"
]},
 "Substance":{"entailment":[
    "cannabis",
    "marijuana",
    "THC",
    "cocaine",
    "heroin",
    "LSD",
    "acid",
    "psilocybin",
    "magic-mushrooms",
    "MDMA",
    "ecstasy",
    "molly",
    "methamphetamine",
    "meth",
    "amphetamine",
    "adderall",
    "methylphenidate",
    "ritalin",
    "ketamine",
    "PCP",
    "phencyclidine",
    "mescaline",
    "peyote",
    "ayahuasca",
    "DMT",
    "salvia",
    "opium",
    "morphine",
    "oxycodone",
    "oxycotin",
    "fentanyl",
    "benzodiazepines",
    "xanax",
    "valium",
    "alcohol",
    "nicotine",
    "tobacco",
    "caffeine",
    "kratom",
    "synthetic-cannabinoids",
    "bath-salts",
    "inhalants",
    "cigarretes",
    "alcohol"
], "contradiction":[
    "sober",
    "abstinent",
    "clean",
    "drug-free",
    "substance-free",
    "non-user",
    "teetotaler",
    "in-recovery",
    "rehabilitated",
    "non-smoking",
    "alcohol-free",
    "straight-edge",
    "clean-living",
    "addiction-free",
    "detoxified",
    "dry",
    "temperate",
    "non-dependent",
    "no-drug-use",
    "no-substance-use"
]},
 "Treatment":{"entailment":[
    "surgery",
    "surgical-procedure",
    "operation",
    "open-surgery",
    "organ-transplant",
    "bone-marrow-transplant",
    "invasive-procedure",
    "chemotherapy",
    "radiotherapy",
    "radiation-therapy",
    "neurosurgery",
    "laparotomy",
    "thoracotomy",
    "ablation",
    "hysterectomy",
    "gastrectomy",
    "amputation",
    "percutaneous",
    "endoscopic-surgery",
    "stent-placement",
    "intravenous-catheter",
    "peritoneal-dialysis",
    "cardiac-catherization",
    "implants",
    "artificial-heart",
    "ventricular-assist-device"
], "contradiction":[
    "medication",
    "pharmacotherapy",
    "physical-therapy",
    "occupational-therapy",
    "psychotherapy",
    "counseling",
    "acupuncture",
    "massage-therapy",
    "chiropractic",
    "herbal-remedies",
    "homeopathy",
    "dietary-changes",
    "exercise",
    "lifestyle-modification",
    "biofeedback",
    "light-therapy",
    "transcutaneous-electrical-nerve-stimulation",
    "TENS",
    "speech-therapy",
    "rehabilitation",
    "orthotics",
    "support-braces",
    "cold-therapy",
    "heat-therapy",
    "hydrotherapy",
    "ultrasound-therapy"
]},
 "Vaccine":{"entailment":[
    "vaccinated",
    "fully-vaccinated",
    "COVID-19-vaccination",
    "COVID-vaccine",
    "immunized",
    "COVID-immunization",
    "COVID-shot",
    "COVID-injection",
    "Pfizer-vaccine",
    "Comirnaty",
    "Moderna-vaccine",
    "Spikevax",
    "JohnsonandJohnson-vaccine",
    "Janssen-vaccine",
    "AstraZeneca-vaccine",
    "Vaxzevria",
    "Sinopharm-vaccine",
    "Sinovac-vaccine",
    "CoronaVac",
    "Covaxin",
    "Covovax",
    "Novavax-vaccine",
    "COVID-booster-shot",
    "booster-dose",
    "COVID-19-vaccinated",
    "immunized-against-COVID",
    "COVID-protection",
    "pandemic-vaccine"
], "contradiction":[
    "unvaccinated",
    "not-vaccinated",
    "COVID-19-unvaccinated",
    "COVID-unimmunized",
    "not-immunized",
    "without-COVID-vaccine",
    "vaccine-free",
    "vaccine-resistant",
    "vaccine-averse",
    "COVID-19-non-vaccinated",
    "unimmunized",
    "vaccine-refusal",
    "unprotected",
    "not-inoculated",
    "without-COVID-shot",
    "without-COVID-injection",
    "vaccine-exempt",
    "non-vaccinated",
    "not-COVID-vaccinated",
    "vaccine-non-compliant"
]},
 "Violence Or Abuse":{"entailment":[
    "abused",
    "victim-of-violence",
    "survivor-of-abuse",
    "trauma-survivor",
    "domestic-violence-victim",
    "sexually-assaulted",
    "physically-assaulted",
    "emotionally-abused",
    "neglected",
    "exploited",
    "harassed",
    "battered",
    "maltreated",
    "beaten",
    "victim-of-trauma",
    "suffered-abuse",
    "experienced-violence",
    "subjected-to-abuse",
    "abuse-victim",
    "assaulted",
    "intimidated",
    "bullied",
    "coerced",
    "manipulated",
    "emotionally-traumatized",
    "psychologically-abused",
    "financially-exploited",
    "sexually-exploited"
], "contradiction":[
    "unaffected",
    "free-from-abuse",
    "abuse-free",
    "trauma-free",
    "safe",
    "secure",
    "protected",
    "well-adjusted",
    "healthy",
    "stable",
    "non-victim",
    "resilient",
    "not-experienced-abuse",
    "not-subjected-to-violence",
    "emotionally-healthy",
    "unharmed",
    "not-exposed-to-abuse",
    "not-assaulted",
    "not-experienced-trauma",
    "non-traumatized",
    "non-exploited",
    "well-supported",
    "non-victimized"
]}
\newpage
\textbf{S1 B.3} STable 1. Natural Language Inference (NLI) statements by entity type dimension. The statements were crafted to represent meaningful binary distinctions in the data.

\begin{figure}[H]
    \centering
    \includegraphics[width=1\linewidth]{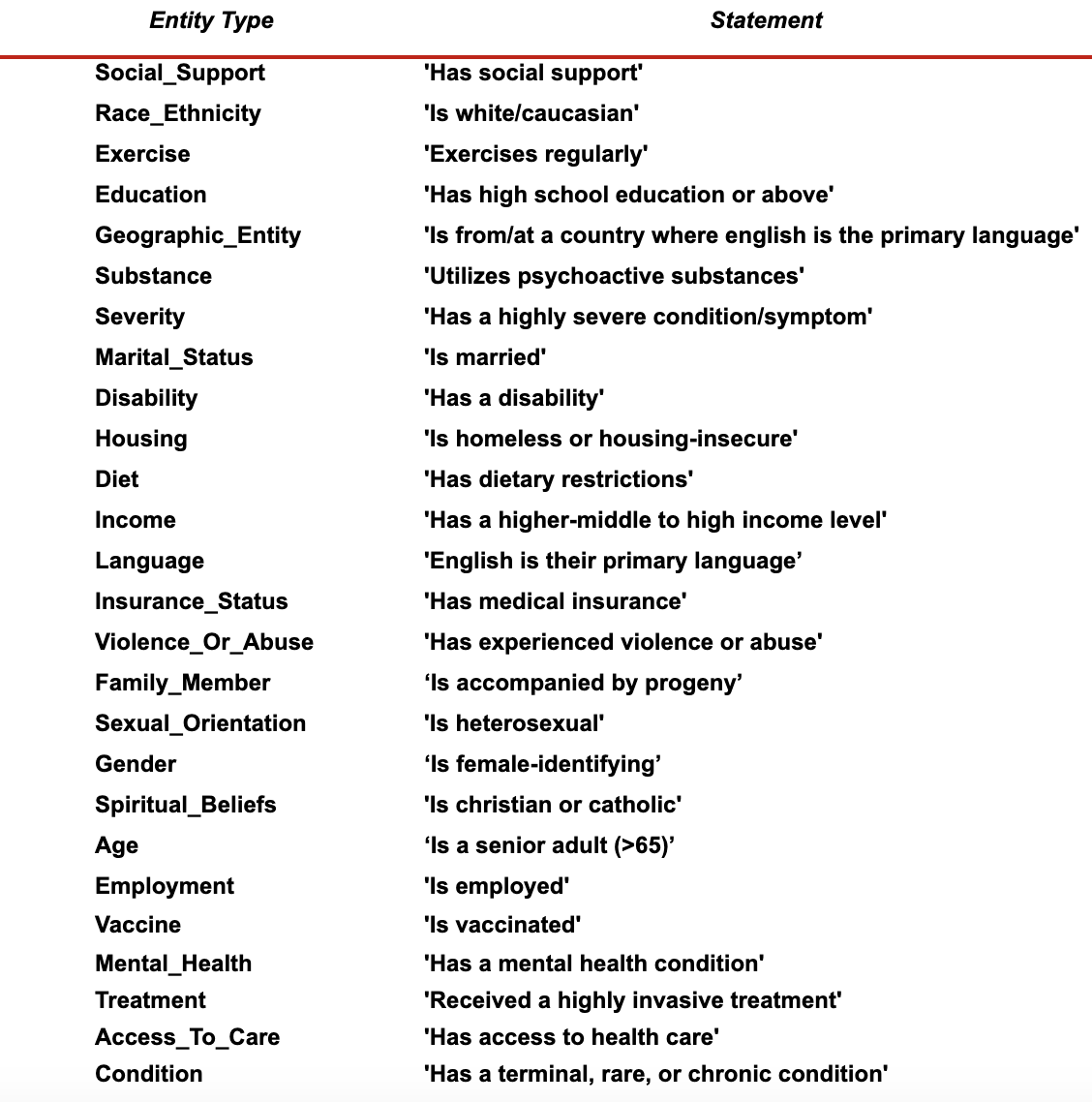}
\end{figure}

\newpage
\textbf{S1 B.4} STable 2. Fine-tuned BERT base uncased performance by entity type on the generalization set

\begin{longtable}{|l|c|c|c|}
\hline
\textbf{Class} & \textbf{Precision} & \textbf{Recall} & \textbf{F1-Score} \\
\hline
\endfirsthead

\hline
\textbf{Class} & \textbf{Precision} & \textbf{Recall} & \textbf{F1-Score} \\
\hline
\endhead

\hline
'O' & 1.00 & 0.69 & 0.82 \\
'B-Age' & 0.36 & 1.00 & 0.53 \\
'I-Age' & 0.71 & 1.00 & 0.83 \\
'B-Gender' & 0.18 & 0.96 & 0.31 \\
'B-Employment' & 0.28 & 0.84 & 0.42 \\
'I-Employment' & 0.48 & 1.00 & 0.65 \\
'B-Condition' & 0.10 & 0.57 & 0.17 \\
'I-Condition' & 0.08 & 0.60 & 0.14 \\
'B-Treatment' & 0.45 & 0.92 & 0.60 \\
'B-Vaccine' & 0.37 & 0.79 & 0.51 \\
'I-Vaccine' & 0.53 & 0.88 & 0.66 \\
'B-Severity' & 0.57 & 0.98 & 0.72 \\
'I-Severity' & 0.69 & 0.94 & 0.79 \\
'B-Geographic Entity' & 0.38 & 0.99 & 0.55 \\
'B-Education' & 0.94 & 0.96 & 0.95 \\
'I-Education' & 0.53 & 1.00 & 0.69 \\
'B-Access To Care' & 0.40 & 0.79 & 0.53 \\
'I-Access To Care' & 0.54 & 0.95 & 0.69 \\
'B-Mental Health' & 0.60 & 0.96 & 0.74 \\
'I-Treatment' & 0.58 & 0.88 & 0.70 \\
'B-Family Member' & 0.68 & 0.73 & 0.70 \\
'I-Geographic Entity' & 0.83 & 1.00 & 0.91 \\
'B-Exercise' & 0.79 & 1.00 & 0.88 \\
'I-Exercise' & 0.46 & 0.99 & 0.63 \\
'I-Family Member' & 0.99 & 0.83 & 0.90 \\
'B-Social Support' & 0.55 & 0.90 & 0.68 \\
'I-Social Support' & 0.82 & 1.00 & 0.90 \\
'B-Housing' & 0.96 & 0.83 & 0.89 \\
'I-Housing' & 0.53 & 1.00 & 0.70 \\
'B-Race Ethnicity' & 0.93 & 1.00 & 0.96 \\
'I-Race Ethnicity' & 0.99 & 1.00 & 0.99 \\
'I-Gender' & 0.34 & 1.00 & 0.51 \\
'I-Mental Health' & 0.85 & 1.00 & 0.92 \\
'B-Substance' & 0.39 & 0.93 & 0.55 \\
'I-Substance' & 0.39 & 0.72 & 0.51 \\
'B-Diet' & 0.25 & 0.66 & 0.36 \\
'B-Marital Status' & 0.87 & 1.00 & 0.93 \\
'B-Income' & 0.66 & 0.90 & 0.76 \\
'I-Income' & 0.98 & 0.92 & 0.95 \\
'B-Language' & 0.90 & 0.94 & 0.92 \\
'B-Violence Or Abuse' & 0.97 & 0.94 & 0.96 \\
'I-Violence Or Abuse' & 0.91 & 0.99 & 0.95 \\
'B-Insurance Status' & 0.31 & 0.47 & 0.37 \\
'I-Insurance Status' & 0.98 & 0.78 & 0.87 \\
'I-Diet' & 0.93 & 0.93 & 0.93 \\
'B-Sexual Orientation' & 0.97 & 0.48 & 0.64 \\
'B-Disability' & 0.47 & 0.97 & 0.63 \\
'I-Sexual Orientation' & 0.93 & 0.69 & 0.79 \\
'I-Disability' & 0.52 & 0.98 & 0.68 \\
'I-Language' & 0.99 & 0.96 & 0.98 \\
'B-Spiritual Beliefs' & 0.80 & 0.95 & 0.87 \\
'I-Spiritual Beliefs' & 0.99 & 0.96 & 0.98 \\
'I-Marital Status' & 0.99 & 1.00 & 0.99 \\
\hline
\end{longtable}

\textbf{S1 B.5} STable 3. Performance of RNN and GRU models

\begin{table}[h]
    \begin{flushleft}
        \fontsize{7}{10}
        \renewcommand{\arraystretch}{1.2} 
        
        \textbf{Optimization Testing Set} \\[1pt] 
        \begin{tabular}{lcccc} 
            \toprule
            \textbf{Model} & \textbf{Macro F1 (all)} & \textbf{Macro AUC (Excl. O)} & \textbf{Macro F1 (OVR)} & \textbf{Macro AUC (OVO)} \\ 
            \midrule
            RNN & 0.96 & 0.96 & 0.99 & 0.99 \\
            GRU & 0.94 & 0.94 & 0.99 & 0.99 \\
            \bottomrule
        \end{tabular} 
        
        \vspace{5pt} 

        \textbf{Generalization Evaluation Set} \\[2pt] 
        \begin{tabular}{lcccc} 
            \toprule
            \textbf{Model} & \textbf{Macro F1 (all)} & \textbf{Macro AUC (Excl. O)} & \textbf{Macro F1 (OVR)} & \textbf{Macro AUC (OVO)} \\ 
            \midrule
            RNN & 0.08 & 0.07 & 0.65 & 0.59 \\
            GRU & 0.09 & 0.08 & 0.69 & 0.66 \\
            \bottomrule
        \end{tabular}

    \end{flushleft}
\end{table}

\section*{Acknowledgements}
Resources used in preparing this research were provided, in part, by the Province of Ontario, the Government of Canada through CIFAR, and companies sponsoring the Vector Institute www.vectorinstitute.ai/partnerships/. This publication was supported by the Canadian Institutes of Health Research (CIHR), Funding Reference Number 192124.

The CAN-TAP-TALENT is funded by the Canadian Institutes of Health Research (CIHR) – FRN 184898. The authors wish to acknowledge the CAN-TAP-TALENT for its role in supporting the completion of this CAN-TAP-TALENT Research Project.
\subsection*{\textbf{CRediT Authorship Contribution Statement}}
\textbf{Juan Andres Medina Florez}: Conceptualization, Methodology, Software, Validation, Formal Analysis, Investigation, Data Curation, Writing - Original Draft, Writing - Review \& Editing, Visualization, Project administration, Funding acquisition. \textbf{Shaina Raza}: Conceptualization, Methodology, Software, Data Curation, Validation, Writing - Review \& Editing. \textbf{Rashida Lynn}: Data Curation. \textbf{Zahra Shakeri}: Supervision, Writing - Review \& Editing. \textbf{Brendan T. Smith}: Supervision, Writing - Review \& Editing, Data Curation. \textbf{Elham Dolatabadi}: Conceptualization, Methodology, Writing - Review \& Editing, Validation, Supervision, Project administration, Funding acquisition.

\subsection*{Declaration of competing interest}
The authors declare that they have no known competing financial interests or personal relationships that could have appeared to influence the work reported in this paper.

\bibliography{References}

\begin{thebibliography}{10}
\bibitem{noauthor_undated-ll}
oncordance between self-reported race/ethnicity and that recorded in a Veteran Affairs electronic medical record;.
\end{thebibliography}

\end{document}